\RequirePackage{fix-cm}
\documentclass{svjour3}                     
\smartqed  
\usepackage{graphicx}
\usepackage{mathptmx}      
\usepackage{comment}
\usepackage[colorlinks,citecolor=blue]{hyperref}
\DeclareGraphicsExtensions{.pdf,.jpg}
\begin{document}

\title{A fast  accurate fine-grain object  detection model based on YOLOv4 deep neural network  
}
\subtitle{}

\titlerunning{Improved object detection model based on YOLOv4}        

\author{Arunabha M. Roy $^{*}$        \and
        Rikhi Bose      \and
        Jayabrata Bhaduri
}


\institute{Arunabha M. Roy ($^{*}$ Corresponding author)  \at
              University of Michigan, Aerospace Engineering, Ann Arbor, MI  48109, U.S.A. \\
              \email{ arunabhr.umich@gmail.com}           
           \and
           Rikhi Bose \at
              Mechanical Engineering, Johns Hopkins University, Baltimore, MD 21218, U.S.A.\\
           \and
           Jayabrata Bhaduri \at
              Capacloud AI, Deep Learning $\&$ Data Science Division, Kolkata, WB 711103, India.
}

\date{Received: date / Accepted: date}

\maketitle

\begin{abstract}
Early identification and prevention of various plant diseases in commercial farms and orchards is a key feature of precision agriculture technology. 
This paper presents a high-performance real-time fine -grain object detection framework that addresses several obstacles in plant disease detection that hinder the performance of traditional methods, such as, dense distribution, irregular morphology, multi-scale object classes, textural similarity, etc. 
The proposed model is built on an improved version of the You Only Look Once (YOLOv4) algorithm. 
The modified network architecture maximizes both detection accuracy and speed by including the DenseNet in the back-bone to optimize feature transfer and reuse, two new residual blocks in backbone and neck enhance feature extraction and reduce computing cost; the Spatial Pyramid Pooling (SPP)  enhances receptive field, and  a modified Path Aggregation Network (PANet) preserves fine-grain localized information and improve feature fusion. 
Additionally, use of the Hard-Swish function as the primary activation improved the model's accuracy due to better nonlinear feature extraction. 
The proposed model is tested in detecting four different diseases in tomato plants under various challenging  environments. 
The model outperforms the existing state-of-the-art detection models in detection accuracy and speed. 
At a detection rate of 70.19 FPS, the proposed model obtained a precision value of $90.33 \%$, F1-score of $93.64 \%$, and a mean average precision ($mAP$) value of $96.29 \%$. 
Current work provides an effective and efficient method for detecting different plant diseases in complex scenarios that can be extended to different fruit and crop detection, generic disease detection, and various automated agricultural detection processes. 

\end{abstract}

\section{Introduction}
\label{intro}

In recent advancements in autonomous agricultural production, informatics is widely utilized commercially to enhance and estimate agricultural yields
\cite{Vougioukas-RAS-2019}. 
Early and accurate disease detection is critical for their prevention and cure, activation of intelligent sprayer systems, and in controlling autonomous pesticide spraying robots in large commercial farms and orchards. 
Exercising such measures result in the reduction of any growth disorders, minimizing pesticide application for pollution-free crop production \cite{Martinelli-ASD-2015}, and consequently, large gains in production. 
With recent advancements of computer vision in precision agriculture technology, crop imaging, and disease detection protocol have become an integral part of collecting crop growth and health monitoring information \cite{Ling-RAS-2019}.
The deep learning techniques have been utilized as an effective tool in agriculture \cite{Kamilaris-CEA-2018} in a wide range of applications, such as but not limited to, crop and fruit classification \cite{Lee-PR-2017}, image segmentation in crops \cite{Dias-CI-2018}, crop detection \cite{Yamamoto-Sensor-2014} etc.

Presently, the majority of studies in crop informatics are geared towards deep learning-based image classification of crop diseases and pests \cite{Zheng-2019,banan2020deep}. 
Only a few studies \cite{Arsenovic-2019,Zhang-IEE-2020} focus on plant diseases detection utilizing region suggestion. 
Compared to traditional machine learning, these models provide superior accuracy, however, due to slow detection speed, these approaches are not suitable for real-time detection in complex environment. 
In that regard, existing disease detection models have limited capability in detecting multi-scale disease distribution, in particular, fine-grain early diseases due to insufficient deep feature extraction. 
In the present work, a novel detection model is proposed based on an advanced computer vision algorithm that demonstrates superior performance in real-time accurate multi-scale plant disease detection in tomatoes.

Compared to the traditional machine learning models, deep learning algorithms have demonstrated higher accuracy in object detection \cite{Han_et_al-2018}. 
The main drawback of these models is that they induce significant error in detecting densely distributed objects with occlusion and overlapping. 
Convolutional neural network (CNN) -based object detection models were developed to address these limitations. 
These models may be classified as two-stage or one-stage detectors \cite{Lin_et_al-2017}. 
One of the well-known two-stage detectors is the Region Convolution Neural Network (RCNN) which includes fast/ faster-RCNN  \cite{Girshick-2015,Ren_et_al-2015} and mask-RCNN \cite{He_et_al-2017}. 
These models are useful in crop and fruit detection, yield, growth evaluation, and automated agricultural management \cite{Bargoti-Underwood-2016}. 
Although faster R-CNN composed of region proposal (RPN) and classification networks leads to a significant drop in detection time, these models can not perform real-time detection with high-resolution images.

More recently, the You Only Look Once (YOLO) algorithm  \cite{Redmon_et_all-2016,Redmon_Farhadi-2017,Redmon_et_all-2018,Bochkovskiy_et_all-2020} was proposed which unifies target classification and localization into a regression problem. 
Since the YOLO does not use RPN, it can directly perform regression to detect targets in an image which significantly increases detection speed. 
The YOLO model have been utilized in various complex detection tasks under challenging scenarios  \cite{wang2020real,martinez2021dynamic,choudhary2021iris}.  
The more recent YOLOv4  has higher detection speed and  performs with better  precision and accuracy compared to YOLOv3 in different real-time object detection applications \cite{Zhu-S-2020,Yu-S-2021} including its application in improving autonomous  detection  \cite{gai2021detection,roy2021deep}. 
However, the original YOLOv4 can provide low detection accuracy with a high number of missed detection and false object predictions due to its insufficient fine-grain feature extraction properties which is essential for early disease detection. 
Additionally, the YOLOv4 incurs high computational cost and longer training time that may not be suitable for in-field mobile computing devices. 
 
To address the aforementioned shortcomings, in the present study, an improved version of the YOLOv4 algorithm \cite{Bochkovskiy_et_all-2020} has been developed for real-time multi-scale disease detection. 
The YOLOv4 algorithm was modified to optimize both detection speed and accuracy. 
The model performance is verified by detecting objects under various challenging environments. 
The contributions of this paper are as follows: in order to improve feature transfer and reuse for small-target detection,  CSPDarkNet53 \cite{Bochkovskiy_et_all-2020} is modified by introducing DenseNet \cite{Huang-IEEE-2017} in the last two feature layers. 
To reduce redundancy and computing cost, the number of network layers is reduced by modifying the convolution blocks. 
We propose a new residual CSP1-$n$ block in the backbone for the improvement of the feature extraction network. 
Moreover, an additional residual CSP2-$n$ block is integrated into modified path aggregation network (PANet) \cite{Liu-IEEE-2018} to preserve fine-grain localized information, feature fusion of multi-scale semantic information, and further reduction of computing cost. 
In addition, the integration of spatial pyramid pooling (SPP) \cite{He-IEEE-2015} block with the backbone of the proposed model enhances receptive field. 
Lastly, the novel Hard-Swish activation \cite{Avenash-2019} has been used to enhance the nonlinear feature learning ability of the network model. 
This was found to increase the accuracy of the model substantially. 
The proposed detection model is tested in detecting four different diseases in tomato plants in various challenging conditions. 

Four diseases are prevalent in tomato plants-- early blight, late blight, Septoria spot, and leaf mold. 
In early phases, due to differences in size, color, cluster density, and other morphological characteristics for these diseases, real-time accurate detection is challenging for traditional methods. 
Moreover, erratic growth pattern, frequent high aspect ratio of the lesions, coexistence of multi-scale object classes, visual similarities and low distinguishable interface  between infected areas and its surroundings, densely populated discreet form of patches, different characteristics in different stages of disease growth, and other critical factors offer additional challenges and difficulties for the object detection models. 
Also, most of the existing models are designed to detect diseases in higher length scales
and  are not efficient/ applicable in real-time detection. 
The model proposed herein outperforms the original YOLOv4 and existing state-of-the-art detection models in terms of both accuracy and speed in detecting all four disease classes. 
Apart from detecting large-scale diseases of different classes in complex scenarios, the modified model performs exceptionally well in detecting diseases at finer grain/ in very small patches. 
Therefore, current work provides an effective method for early detection of different plant diseases in complex scenarios which may be used in real-time detection of diseases in fields from portable mobile computing devices.

The paper is organized as follows: Section 2 introduces the YOLO framework; Section 3 describes the proposed network algorithm;   object detection preliminaries including performance metrics, bounding box regression, and loss functions have been described in Section 4; Section 5 demonstrates the proposed model's superiority comparing its performance with state-of-the-art object detection models, specifically, original YOLOv3 and YOLOv4 models in tomato plane disease detection problem under various complex scenarios. The model's real-time detection performance in detecting four diseases in tomato plants is shown in Section 6. Several performance measures defined in Section 4 were utilized in sections 5 and 6. Finally, the conclusions and prospects of the current work have been discussed in section 7.

\section{YOLOv4 network structure}
\label{sec:2}

\begin{figure}
	\noindent
	\centering
	\includegraphics[width=1.1\linewidth]{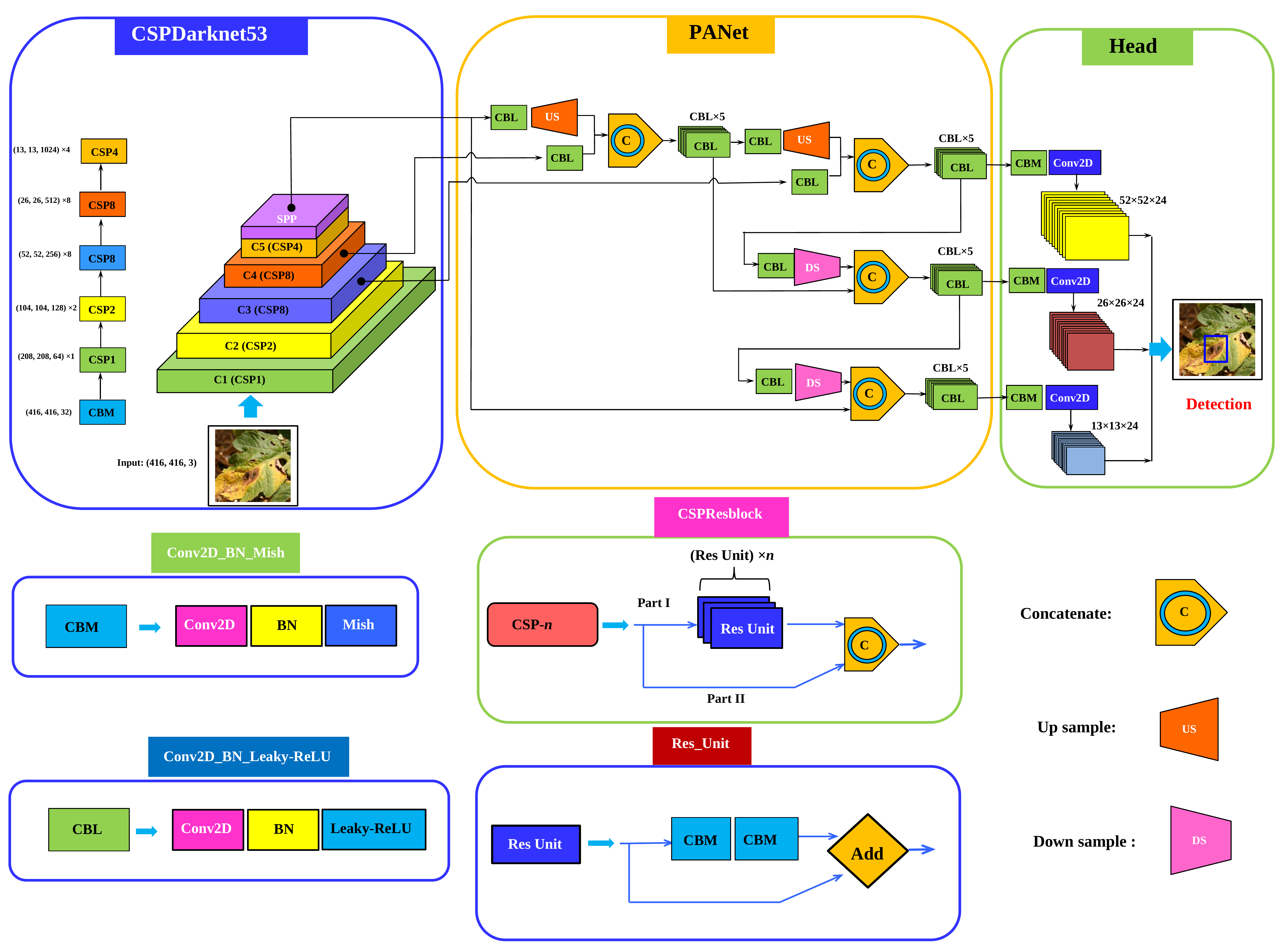}
	\caption{\label{Fig-1} Schematic of the YOLOv4 network architecture  consisting of CSPDarknet53 as the backbone,  PANet as the neck with a regular YOLOv3 head.}
\end{figure}

In the present work, an improved YOLOV4 algorithm \cite{Bochkovskiy_et_all-2020} is implemented for the purpose of plant disease detection. 
The YOLOv4 is a  high-precision single-stage object detection model which transforms the object detection task into a regression problem by generating bounding box coordinates and assigning probabilities to each class. 
The YOLOv4 is an improved version of the original YOLO algorithm \cite{Redmon_et_all-2016} and it's offsprings, YOLOv2 \cite{Redmon_Farhadi-2017}, and YOLOv3 \cite{Redmon_et_all-2018} in terms of detection speed and accuracy. 
As shown in Fig. \ref{Fig-1}, the complete  network structure consists of  three parts : a backbone for feature extraction, the neck for semantic representation of extracted features, and the head for prediction. 

\begin{figure}
	\noindent
	\centering
	\includegraphics[width=0.9\linewidth]{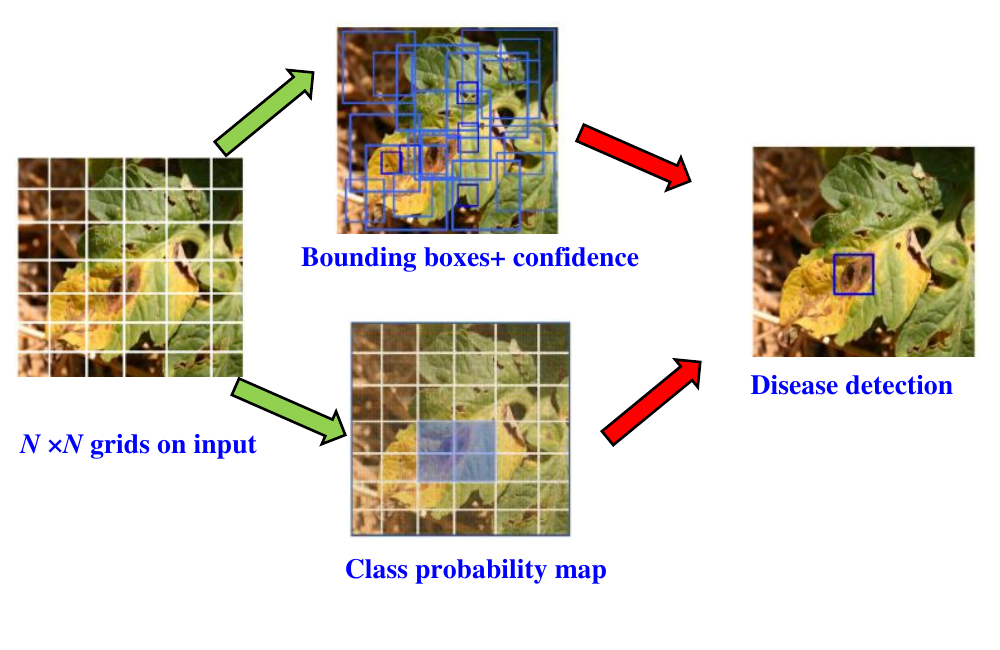}
	\caption{\label{Fig-2}  Schematic of YOLOv4 object detection algorithm for disease detection.}
\end{figure}

In the network architecture, the residual module is integrated into ResNet network structure \cite{Wu-PR-2019} to obtain Darknet53. 
For further improvement of the network performance, cross-stage partial network (CSPNet) \cite{Wang-CVPR-2020} is combined considering its superior learning capability to form CSPDarkNet53 \cite{Bochkovskiy_et_all-2020}.
Different feature layer's information is inputted into the residual module which provides the higher-level feature maps as the output. 
This results in a significant decrease in network parameters, at the same time improving the residual feature information, and enhancing feature learning capability compared to the ResNet network. 
In the original YOLOv4  backbone, the SPP block \cite{He-IEEE-2015} was integrated with CSPDarknet53 linked to the PANet \cite{Liu-IEEE-2018}, replacing the feature pyramid networks (FPN) \cite{Lin-IEEE-2017} used in other variants of the YOLO. 
This resulted in a significant increase of the receptive field. 
The SPP applies an effective strategy for detecting objects of different scales. 
At first, the inputted feature layer is convoluted in SPP. 
Afterward, a maximum pool can be applied by pooling cores of a maximum of four different sizes. 
Pooled feature information obtained from the SPP is then concatenated and further convoluted which significantly increases the receptive field of the detection network. 
Obtained feature information fields from the backbone and the SPP is convoluted and then up-sampled in PANet resulting in twice the size of the inputted feature layer. 
To extract additional semantic features, the feature layers obtained from the CSPDarknet53 are concatenated after convolution, then up-sampled followed by down-sampling which is stacked with the remaining feature layers for enhancing the feature fusion process, as shown in Fig. \ref{Fig-1}. 
Thus, the neck is leveraged in backbones for the extraction of rich semantic features that are used for accurate predictions. 
Finally, for the specific inputted image size, the YOLOv4 model can predict bounding boxes at the detection head at three different scales. 
In the first step, the inputted image is discretized into $N \times N$ equally spaced grids. 
The model generates $B$ predictive bounding boxes and a corresponding confidence score if the target belongs within a grid-cell. 
The best bounding box prediction from each of these scales is filtered by non-maximum suppression (NMS) \cite{Ren_et_al-2015} algorithm before the final bounding box can be obtained. 
The prediction process is shown in Fig. \ref{Fig-2}. 
In order to help the model learn various types of distribution of a given image in challenging circumstances, in particular, noise, complex backgrounds etc., YOLOv4 introduces CutMix \cite{Yun-IEEE-2017}, mosaic augmentations \cite{Bochkovskiy_et_all-2020}, and self-adversarial training (SAT) \cite{Bochkovskiy_et_all-2020}  methods to expand the dataset. 
Additionally, drop block regularization \cite{Ghiasi-NIPS-2018} for learning spatially discriminating features and class label smoothing \cite{Bochkovskiy_et_all-2020} for better generalization of a dataset can be employed. 

Although the aforementioned techniques can effectively improve the detection accuracy of the model, the plant disease detection task faces several specific challenges due to complex environments, in particular, densely populated fine-grain disease, irregular geometric morphology of infected areas, coexistence of multi-scale infected lesions, similarity of texture of affected areas and surroundings, varying lightening conditions, overlapping and occlusion, etc. 
Thus original YOLOv4 can provide low detection accuracy that may lead to a high number of missed detection as well as false object prediction due to insufficient fine-grain feature extraction for the multi-scale disease detection problem. 
Additionally, the YOLOv4 incurs high computational cost and longer training time that may not be suitable for on-field mobile devices. 

\section{The proposed model network structure}
\label{sec:3}

\begin{figure}
	\noindent
	\centering
	\includegraphics[width=1.1\linewidth]{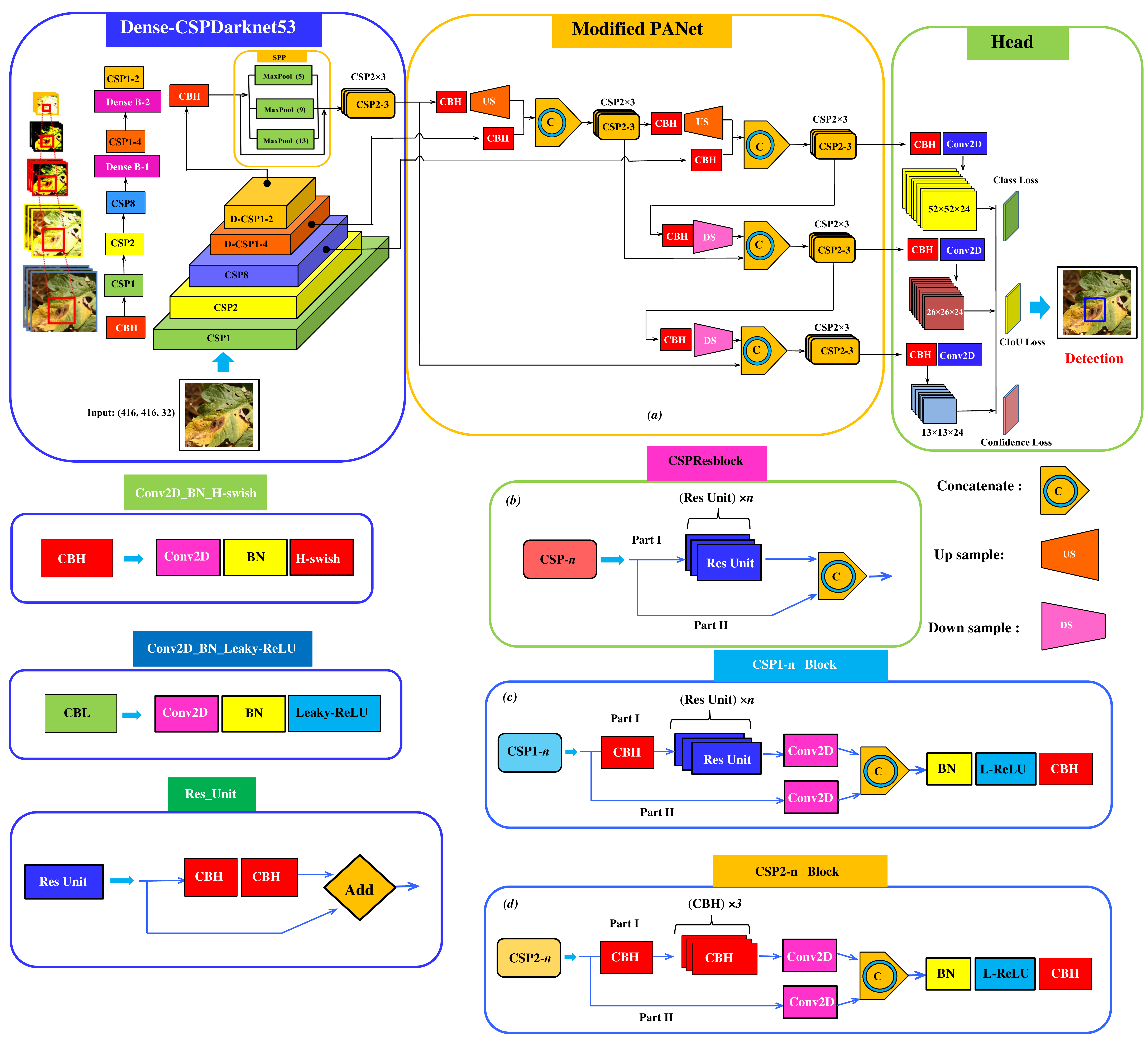}
	\caption{\label{Fig-3}  Schematic of (a) the proposed network architecture for plant disease detection  consisting of Dense-CSPDarknet53 integrating SPP as the backbone, modified PANet as a neck with a regular YOLOv3 head; (b) CSP-$n$; (c) CSP1-$n$; (d) CSP2-$n$ blocks.}
\end{figure}

In order to resolve the aforementioned issues related to real-time disease detection procedure, in the present study, the state-of-the-art YOLOv4 algorithm is improved and optimized for accurate predictions of fine-grain image multi-attribute detection in complex background environments. 
In the later sections, efficacy of the model is demonstrated by detecting different tomato plant diseases at real-time detection speeds in complex backgrounds. 
The complete schematic of the improved YOLOv4 network architecture is shown in Fig. \ref{Fig-3} and each modification is briefly discussed in this section. 
The proposed modifications include the proper selection of activation functions for the backbone and the neck for better accuracy, inclusion of the DenseNet \cite{Huang-IEEE-2017} transitional block in front of the residual blocks of the original CSPDarknet53, two new residual blocks in the backbone and the neck to enhance feature extraction network and to reduce the computing cost, integration of the SPP  \cite{He-IEEE-2015} block, and the implantation of the modified PANet \cite{Liu-IEEE-2018} 
in the neck part of the network to preserve fine-grain localized information. 
The original YOLOv3 head is used as the detection head. 
With an  inputted image size of $416\times416\times3$, the proposed model can predict bounding boxes at the detection head at three different scales, $52\times52\times24$, $26\times26\times24$, and $13\times13\times24$. 
It is found that with the aforementioned modifications, the model outperforms other state-of-the-art detection models, specifically at detecting fine-grain tomato plant diseases. 

\subsection{Improvement of feature extraction network in the backbone}
\label{sec:3a}

The residual model in the CSPDarknet53 helps the network to learn more expressive features at the same time reducing the number of trainable parameters to make it faster for real-time detection. 
In the original YOLOv4 model, the residual unit (Res-unit) carries out $1\times1$ convolutions, followed by $3\times3$ convolutions, and then weights the two outputs containing extracted feature information at the end. 
In the CSPDarknet53 network, feature layers of the inputted images are continuously down-sampled via the convolution operations to extract fine-grained rich semantic information. 
As the last three layers contain relatively higher semantic information, these are passed to the SPP and the PANet. 
The last feature layers contain the finest feature information, and is connected to the SPP. 
The other two layers are integrated into the PANet as shown in Fig. \ref{Fig-3}. 
Although the residual module in the YOLOv4 reduces computational cost, this further reduces computational memory requirement for high-resolution real-time detection. 
Therefore, a new residual block, CSP1-$n$ \cite{Zhu-S-2020} ($n$ is the number of residual weighting operations) is proposed in the CSPDarkNet53 network structure (see Fig. \ref{Fig-3}) to improve detection speed and performance.  
In the proposed CSP1-$n$ residual block, the input features are divided into two parts. 
The first part of the residual block acts as the trunk which performs $1\times1$ convolution followed by another $1\times1$ convolution to adjust the channel after entering the main residual unit as in Fig. \ref{Fig-3}(c). 
To enhance feature extraction even further, it then performs $3\times3$ convolution, whereas, the second part acts as a residual edge for the convolution. 
At the end of the CSP1-$n$ block, these two parts are concatenated resulting in additional feature layer information. 
In the present study, CSP1-$n$ modules replace the CSP8 and CSP4 in an improved backbone. 
Finally, $1\times1$ convolution performed to integrate the channel after stacking.  Implementation of the CSP1-$n$ modules in the modified CSPDarknet53 significantly improves the detection accuracy for the feature dataset used herein. 

\subsection{Implementation of the Hard-swish activation for better accuracy}
\label{sec:3b}

One of the important aspects of developing an object detection model is to select appropriate activation function for better accuracy and performance \cite{Ramachandran-2018}. 
Activation functions can be characterized by properties such as, derivative, monotonic behavior, etc \cite{Eger-2019}. 
In this regard, Leaky Rectified Linear Unit (Leaky-ReLU) \cite{Maas-2013}, Mish \cite{Misra-2019} are widely used activations in dense object detection models. 
However, using the Swish function \cite{Ramachandran-2018} as a drop-in replacement for ReLU demonstrates significant improvement of the  neural network performance \cite{Ramachandran-2018,Elfwing-2017,Hendrycks-2016}. 
The Swish function is expressed as, $swish(x)= x. \sigma(x)$. 
Due to presence of sigmoid function $\sigma(x)$, it increases computational cost. 
Therefore, the Hard-swish activation function \cite{Avenash-2019}, where $\sigma(x)$ in the Swish function is replaced with its piece-wise linear hard analog, $ReLU6(x+3)$, is used instead. 
The function is, 
\begin{equation}
	H-swish(x)= x. \frac{ReLU(x+3)}{6}
	\label{E-2}
\end{equation}

Due to H-swish's unique property of non-monotonicity, it can improve the performance of the detection model for different datasets. 
Additionally, due to H-swish is bounded below and its property of unboundedness, it helps remove the saturation problem of the output neurons and improve network regularization. 
Moreover, it is computationally faster than Swish and beneficial for training as it helps to learn more expressive features that are more robust to noise \cite{Avenash-2019}. 
Hard-swish activation is used in different object detection algorithms which substantially reduces the number of memory accesses by the model \cite{Howard-IEEE-2019,Yu-Sensors-2021}. 
Hard-Swish function is used herein as the primary activation in both the backbone and the neck with significant accuracy gain on the dataset under consideration. 
Moreover, the detection speed is increased and computational cost is substantially reduced (see Section 5.1.1). 

\subsection{Implementation of the DenseNet for better feature transfer and reuse}
\label{sec:3c}

During object detection, the YOLOv4  algorithm reduces the feature maps during training. 
Due to several steps of convolution and down-sampling procedures, important feature information of the training sample can get lost during transmission. 
To preserve important feature maps and to reuse critical feature information more efficiently, the DenseNet framework \cite{Huang-IEEE-2017} is proposed where each layer is connected to other layers feeding forward. 
The main advantage of this framework is that the $n$-th layer is able to receive required feature information $X_n$ from all previous layers as inputs. 

\begin{equation}
	X_n= H_n[X_0, X_1,...,X_{n-1}]
	\label{E-3}
\end{equation}

\begin{figure}
	\noindent
	\centering
	\includegraphics[width=1.1\linewidth]{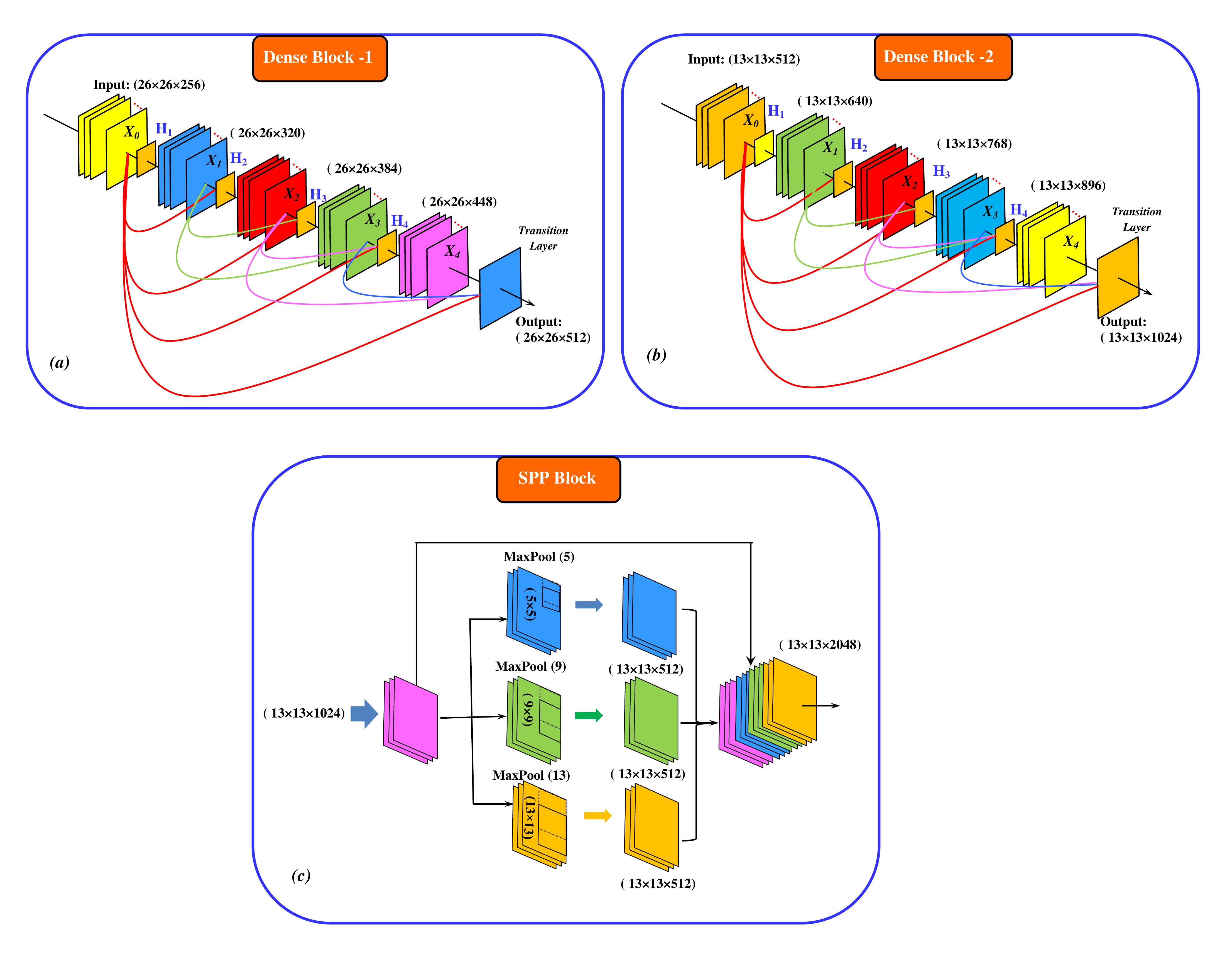}
	\caption{\label{Fig-4} Schematic of (a) dense block-1; (b) dense block-2 in Dense-CSPDarknet53; (c) SPP block and  corresponding network parameters of the proposed disease detection model.}
\end{figure}

\noindent Here, $H_n$ is the spliced feature map function for layer $n$;  $[X_0, X_1,...,X_{n-1}]$ is the feature map of layers $X_0, X_1,...,X_{n-1}$. 
Such formulation allows the DenseNet to reduce the number of parameters, enhance
feature propagation and facilitate feature reuse. 
Due to the complexity of the image dataset, in particular, densely populated distribution and coexistence of multi-scale disease classes, it is critical to use the dense block to facilitate better feature transfer and gradient propagation throughout the network.  Additionally, it may mitigate over-fitting to some degree. 
In the proposed model, the last two residual blocks, CSP8 and CSP4 in the original CSPDarknet53 are modified to Dense-CSP1-4 and Dense-CSP1-2 by adding dense connection blocks to enhance feature propagation. 
Additionally, the redundant feature operations is reduced and the calculation speed is increased by removing the CSP-$n$ blocks. 
The schematic of the proposed dense blocks and corresponding network parameters are shown in Fig. \ref{Fig-4}(a, b). 
It is evident that the network structure is improved by replacing  $26\times26$ and $13\times13$ down-sampling layers by the DenseNet structure. 
In the dense block-1, transfer feature map function $H_1$ nonlinearly transforms the $X_0, X_1,...,X_{n-1}$ layers, where each layer $X_i$ is comprised of $64$ feature layers each with resolution $26\times26$ pixels as shown in Fig. \ref{Fig-4}(a). 
The first dense block before the CSP1-4 performs feature propagation and layer splicing on the layers with $26\times26$ resolution which results in the final forward propagating feature layer of size $26\times26\times512$. 
Similarly, feature propagation and layer splicing are performed on the layers with $13\times13$ resolution which result in the final forward propagating feature layer of resolution $13\times13\times1024$ by the second dense block before the CSP1-2 as shown in Fig. \ref{Fig-4}-(b). 
The Dense-CSPDarknet53 configuration ensures that later feature layers obtain features from the previous layers during training when inputted images are transferred to the lower resolution layers of the network reducing the feature loss. 
Moreover, different low-resolution convolution layers can reuse the feature between them  which increases the feature usage rate. 

\subsection{Enhancement of the receptive field}
\label{sec:3d}

To enhance the receptive field and separate important context features during object detection, an SPP block \cite{He-IEEE-2015} is tightly integrated with the last residual block (CSP1-2) of the Dense-CSPDarknet53 backbone structure as shown in Fig. \ref{Fig-3}. 
The SPP applies an effective strategy in detecting objects at different length scales. 
It replaces the pooling layer (after the last convolutional layer) with a spatial pyramid-type pooling layer. 
In the proposed modification, the SPP retains the spatial dimension of the output, as a  maximum pooling is applied to a sliding kernel of sizes $5 \times5$, $9 \times9$, and $13 \times13$ with a stride of 1 as shown in Fig. \ref{Fig-4}-(c). 
A relatively large $13 \times13$ max-pooling effectively increases the receptive field of the backbone. 

\subsection{Modified PANet to preserve fine-grain localized information}
\label{sec:3e}

In the object detection neural network model developed herein, earlier layers extract localized texture and pattern information to build up the semantic information needed in the later layers. 
However, with increasing layers of the residual blocks, interconnectivity among layers become more complex, especially due to the dense connection block where each layer is connected to all previous layers. 
This requires fine-tuning of the localized information. 
In order to address this issue, the PANet \cite{Liu-IEEE-2018} is used in the neck part of the proposed model which shortens the path of high and low fusion for the multi-scale feature pyramid map. 
The PANet fuses information from all layers using element-wise max operation and has more flexible ROI pooling compared to the FPN \cite{Lin-IEEE-2017}. 
To disseminate information at lower levels, a bottom-up path augmentation is used in the PANet as shown in Fig. \ref{Fig-3}. 
The CSP2-$n$ module \cite{Zhu-S-2020} is added to the PANet which divides the basic feature layer into two parts and reduces the use of repeated gradient information through cross-stage operation as shown in Fig. \ref{Fig-3}-(d). 
This further improves multi-scale local feature fusion with global feature information. 
The introduction of the CSP2-$n$ improves the feature extraction flow which leads to a notable increase in detection accuracy and speed (see Section 5.1.2).

\begin{figure}
	\noindent
	\centering
	\includegraphics[width=1\linewidth]{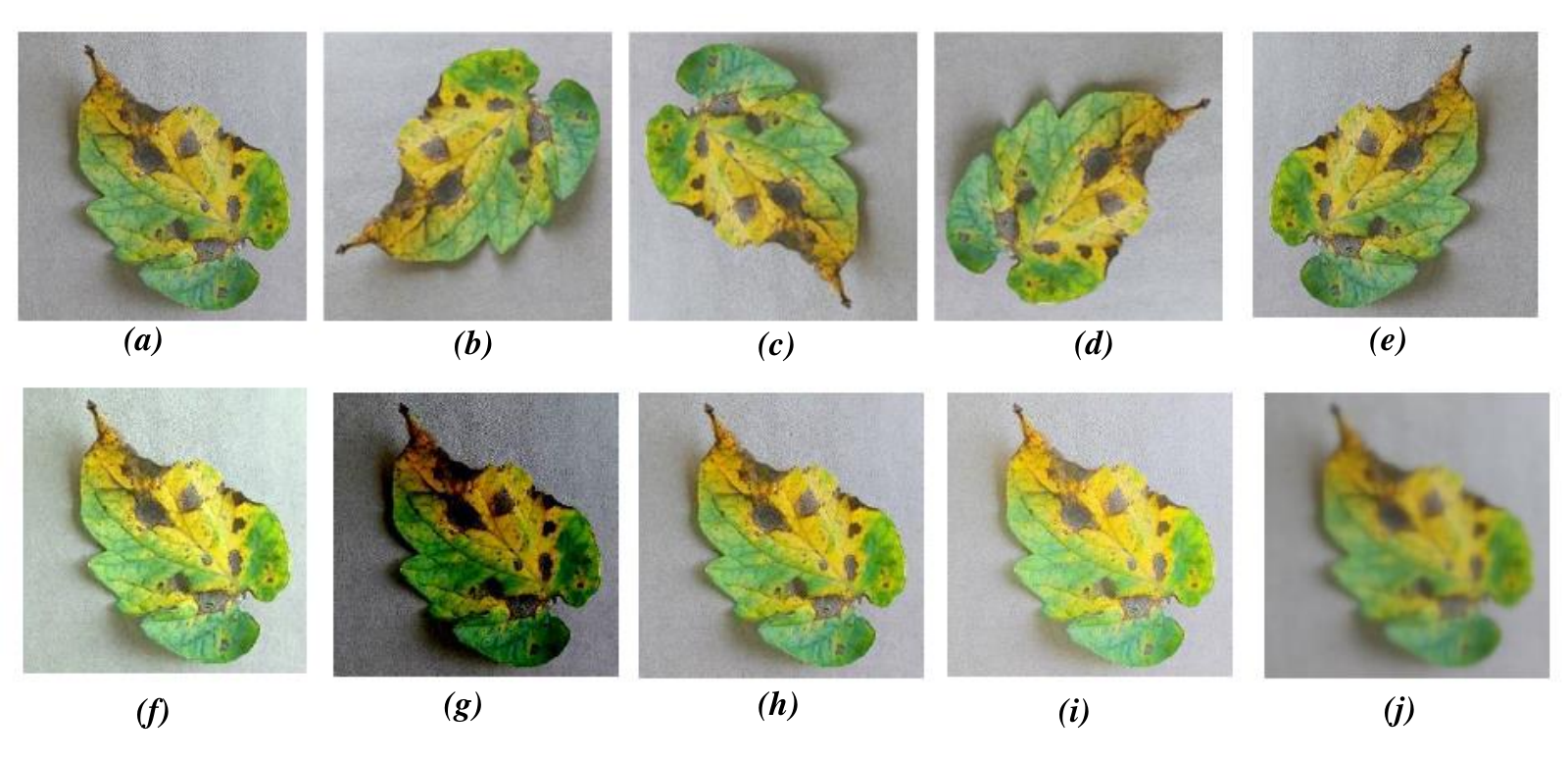}
	\caption{\label{Fig-5} Different image augmentation methods: (a) original image, (b) $90^o$ ACW
		rotation, (c) $180^o$ ACW rotation, (d) $270^o$ ACW rotation, (e) horizontal
		mirror projection, (f) colour balancing, (g-i) brightness transformation, and (j)
		blur processing.}
\end{figure}

Apart from the above mentioned modifications of the model architecture, dropout in the feature map  \cite{Srivastava-2014}, CIoU loss function \cite{Zheng_et_all-2020},  Cross mini Batch Normalization \cite{Yao-2020}, cosine annealing scheduler \cite{Loshchilov-2016}, and dropblock regularization \cite{Ghiasi-2018}, are utilized to improve the performance of the proposed model. 
Additionally, data augmentation procedures, such as, rotation, mirror projection, color balancing,  brightness transformation, blur processing are employed to increase the variability of inputted images obtained from different environments to augment robustness of the detection model as depicted in Fig. \ref{Fig-5}. 

\section{Preliminaries of an object detection problem}
\label{sec:4}

In this section, some relevant detail associated with the object detection problem formulation, such as, the objective of the model, preliminaries of implementation including the loss functions relevant to the object detection problem, and model performance metrics are briefly described. 

\subsection{Bounding box regression}
\label{sec:4a}

In dense object detection models, bounding box regression is a widely applied approach to predict the localization boxes on input images. 
In this regard, a scale-invariant evaluation metric called intersection over union (IoU) was proposed to evaluate the accuracy of target object  detection,

\begin{equation}
	IoU=\frac{\mathbf { B}\,\,\cap\,\,\mathbf {B}_{gt}}{\mathbf { B}\,\,\cup\,\,\mathbf {B}_{gt}}
	\label{E-11}
\end{equation}

where, $\mathbf { B}\,\,\cap\,\,\mathbf {B}_{gt}$ and $\mathbf { B}\,\,\cup\,\,\mathbf {B}_{gt}$ are defined as the intersection and union  between areas of the predicted box $\mathbf {B}$ and the ground truth bounding box $\mathbf {B}_{gt}$ of the object, respectively. 
However, $IoU$ loss only considers the overlapping bounding boxes. 
The non-overlapping cases are not taken into account. 
Additionally,  traditional $IoU$ loss has the limitation on gradient disappearance in case of an absence of intersection between target and prediction boxes.
To circumvent these issues, generalized $IoU$ ($GIoU$) \cite{Rezatofighi_et_all-2019} was proposed which considers the shape, area, and orientation of the overlapping bounding boxes. 
The $GIoU$ loss is expressed as, 
\begin{equation}
	L_{GIoU}=1-IoU+ \frac{|\mathbf {C}-\mathbf {B}\,\,\cup\,\,\mathbf {B}_{gt}|}{\mathbf {C}}
	\label{E-12}
\end{equation}
Where $\mathbf {C}$ is the  smallest convex hull between $\mathbf {B}$ and $\mathbf {B}_{gt}$.
However, $GIoU$ is not computationally cost effective. 
For better performance, distance-$IoU$ ($DIoU$) \cite{Zheng_et_all-2020} loss was introduced which measures the proximity of the target and prediction boxes by introducing a metric parameter in predicting boundary box regression. 
The expression for the $DIoU$ loss is, 
\begin{equation}
	L_{DIoU}= 1-IoU+\frac{\rho^2(\mathbf {b, b_{gt}})}{c^2}.
	\label{E-13}
\end{equation}
\noindent Here, $\mathbf {b}$ and  $\mathbf {b_{gt}}$ are the centroids of $\mathbf {B}$ and   $\mathbf {B}_{gt}$, respectively; $d:=\rho(\mathbf {b, b_{gt}})$ is the distance between central points of $\mathbf {B}$ and $\mathbf {B}_{gt}$, and  $c$ is the length of the diagonal of the smallest enclosing box covering the two boxes as shown in Fig. \ref{Fig-6}-(b).  
The penalty term $\frac{\rho^2(\mathbf {b, b_{gt}})}{c^2}$ in Eq. \ref{E-13} minimizes the normalized distance between central points of the two bounding boxes, leading to faster convergence than the $GIoU$ loss. 
Subsequently, complete $IoU$ ($CIoU$) \cite{Zheng_et_all-2020}, an extension of the $DIoU$ loss,  was introduced in YOLOv4 to better the accuracy and convergence speed for the target bounding box prediction process. 
$CIoU$ loss simultaneously considers three geometric metrics that are typically ignored: overlapping area, the distance between centers, and the aspect ratio in the bounding box regression in object detection \cite{Zheng_et_all-2020}. 
$CIoU$ loss was formulated incorporating consistency of the aspect ratio parameter, $v$, and a positive trade off parameter, $\alpha$ based on $DIoU$ loss which is expressed as, 
\begin{eqnarray}
	\label{E-14}
	&&L_{CIoU}= 1-IoU+\frac{\rho^2(\mathbf {b, b_{gt}})}{c^2}+ \alpha v.
	\\
	\label{E-15}
	&& v= \frac{4}{\pi^2}  \left(tan^{-1}\,\frac{w_{gt}}{h_{gt}}-tan^{-1}\,\frac{w}{h}\right)^2;\quad \alpha= \frac{v}{(1-IoU)+v'}
\end{eqnarray}
Here, $w_{gt}$, $w$ and $h_{gt}$, $h$ are the widths and heights of the ground truth and prediction bounding boxes, respectively as shown in Fig. \ref{Fig-6} -(b). 
In Eq. \ref{E-15}, $v\rightarrow0$ with increase in $w/h$. 
For consistent predictions, $w/h$ is a chosen parameter for a specific YOLOv4 object detection model. 

\subsection{Confidence score}
\label{sec:4b}

During object detection, when the center of the target-class ground truth falls within a specified  grid-cell, it detects the target as an object of a particular class. 
Each grid predicts $B$ bounding boxes with confidence scores and corresponding $C$ class conditional probabilities for each target class. 
The confidence scores can be expressed as, 
\begin{equation}
	confidence=p_r(object)\times IoU^{truth}_{pred} \, \vee\,p_r(object) \in [0, 1]
	\label{E-4}
\end{equation} 
During the object detection process, $p_r(object)=1$ is prescribed in the framework when the target class falls within the YOLO grid-cell, or otherwise, $p_r(object)=0$. 
Coalescing between the reference and the predicted bounding box is quantified by  $IoU^{truth}_{pred}$. 
The value of $p_r(object)$ indicates the accuracy of the bounding-box prediction when the target class is detected within the cell. 
%
\begin{figure}
	\noindent
	\centering
	\includegraphics[width=1\linewidth]{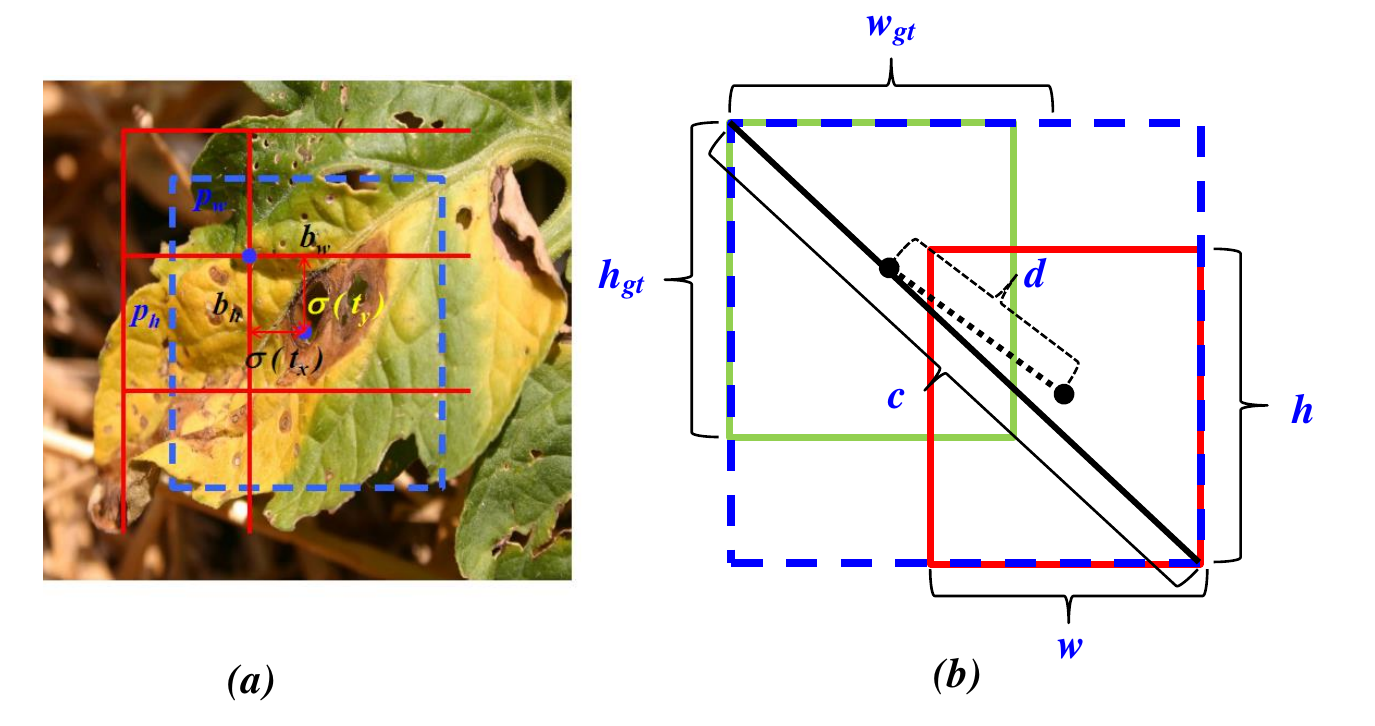}
	\caption{\label{Fig-6}
		(a) Schematic of offset regression for target bounding box prediction process; (b) schematic  of CIoU loss for bounding box regression in  YOLOv4 object detection algorithm.}
\end{figure} 

The prediction process for the target bounding box is shown in Fig. \ref{Fig-6}. 
The dashed box in Fig. \ref{Fig-6}-(a) is the initial bounding box. 
The relationship between the initial bounding box to the predicted bounding box can be  expressed as, 
\begin{eqnarray}
	\label{E-5}
	&&b_x=\sigma(t_x)+c_x; \quad \quad b_y=\sigma(t_y)+c_y.
	\\
	\label{E-6}
	&& b_w=p_we^{t_w};\quad \quad \quad\,\,\,\, b_h=p_he^{t_h}.
\end{eqnarray}
\noindent where, $(b_x, b_y)$ and $(b_w, b_h)$  are the centroid and size of the predicted bounding box; $(c_x, c_y)$ and $(p_w, p_h)$ are the centroid and size of the bounding box on the feature map; $(t_x, t_y)$ and $(t_w, t_h)$ represent the  center offset of the bounding box from the network prediction and corresponding scaling size. 

\subsection{Loss function}
\label{sec:4c}

The loss or cost function ($\xi$) is important as it dictates the performance of the trained model. 
The loss function $\Delta_l$ for an object detection task such as in YOLO can be defined as,
\begin{equation}
	\Delta_l= \Theta_{cor}+ \Theta_{IoU}+\Theta_{cl}.
	\label{E-16}
\end{equation}
$\Delta_l$ consists of mainly three types of error components. 
The coordinate prediction error, $\Theta_{cor}$ is defined as: 
\begin{equation}
	\Theta_{cor}= \kappa_{cor}\sum\limits_{i=1}^{N^2} \sum\limits_{j=1}^B \delta_{ij}^{obj}[(x_i-\bar{x_i})^2+(y_i-\bar{y_i})^2]+
	\kappa_{cor}\sum\limits_{i=1}^{N^2} \sum\limits_{j=1}^B \delta_{ij}^{obj}[(w_i-\bar{w_i})^2+(h_i-\bar{h_i})^2]
	\label{E-17}
\end{equation}
Here, $N^2$ is the total number of grid points in the inputted image, $\kappa_{cor}$ corresponds to the weight associated with $\Theta_{cor}$, $B$ is the total number of bounding boxes associated with each grid, $(x_i, y_i)$ are the true center coordinates of the object, $(\bar{x_i},\bar{y_i})$ are the center coordinates of the predicted bounding box; $(w_i, h_i)$ and $(\bar{w_i},\bar{h_i})$ are the width and height of the truth and the predicted bounding boxes, respectively. 
The function $\delta_{ij}^{obj}=1$ if the target class lies in the bounding box $j$ generated by grid $i$, else $\delta_{ij}^{obj}=0$. 
The $IoU$ error term, $\Theta_{IoU}$ is given as follows.  
\begin{equation}
	\Theta_{IoU}= \sum\limits_{i=1}^{N^2} \sum\limits_{j=1}^B \delta_{ij}^{obj}(C_i-\bar{C_i})^2+
	\kappa_{nb}\sum\limits_{i=1}^{N^2} \sum\limits_{j=1}^B \delta_{ij}^{obj}(C_i-\bar{C_i})^2
	\label{E-18}
\end{equation}
Here, $\kappa_{nb}$ is a parameter corresponding to weight associated with $\Theta_{IoU}$, $C_i$ is the true confidence in object detection, and $\bar{C_i}$ is the confidence score of the prediction. 
Lastly, the classification error term $\Theta_{cl}$ is given by, 
\begin{equation}
	\Theta_{cl}= \sum\limits_{i=1}^{N^2} \sum\limits_{j=1}^B \delta_{ij}^{obj}
	\sum\limits_{c \in classes}
	(p_i(c)-\bar{p_i}(c))^2
	\label{E-19}
\end{equation}
In Eq. \ref{E-19}, $c$ is the class associated with target detection; $p_i(c)$ is the true probability of detecting the object of class $c$ in grid $i$; $\bar{p_i}(c)$ is the probability score from the prediction. 
$\Theta_{cl}$ for a particular grid $i$ is obtained from the sum of classification errors due to all classified objects residing in that grid-cell. 
\\
\\

\subsection{Evaluation metrics}
\label{sec:4d}

Some traditional evaluation parameters, such as, precision-recall ($PR$) curve, F-1 score, average precision (AP), or mean average precision (mAP) can be used \cite{Goutte-2005} to measure the performance of an object detection model. 

For binary classification, sample data can be classified into four different categories: true positive (TP), false positive (FP), true negative (TN), and false negative (FN), based on  the true class and the model predicted class of the  target object. 
From the quantities defined above, Precision ($P$) and Recall ($R$) can be defined  as 
\begin{equation}
	P= \frac{TP}{(TP+FP)}; \quad\quad\quad\quad R=\frac{TP}{(TP+FN)}
	\label{E-7}
\end{equation}

From Eq. \ref{E-4}, one can infer that $P$ represents prediction results of relevant instances. 
$R$ refers to correctly classified results out of the total relevant instances. 
Both  higher $P$ and $R$ indicate lower FN value. 
For a particular set of training sample,  the precision-recall curve
($P-R$ curve) can be  constructed from the precision (in the ordinate) and recall data (in the abscissa) for a particular classifier. 
From the relation between $P$ and $R$, F1-score is defined which indicates the degree of precision of the object detection model. 
From Eq. \ref{E-4}, the F1- score can be obtained as,
\begin{equation}
	F_1= \frac{2PR}{(P+R)}.
	\label{E-8}
\end{equation}

$AP$ is equal to  the area under the $P-R$ curve. 
\begin{equation}
	AP= \int_0^1 \! P (R) \, \mathrm{d}R.
	\label{E-9}
\end{equation}
High AP corresponding to a large area under the PR curve indicates accurate prediction of an object class by a detection model. 
Additionally, $AP_{50:95}$ is the average precision over the range of IoU=$0.50:0.05:0.95$; $AP_{50}$ and $AP_{75}$ are $AP$s  at IoU thresholds $50\%$ and $75\% $, respectively; $AP_{S}$, $AP_{M}$, and $AP_{L}$ correspond to the detection accuracy of small, medium, and large objects, respectively, for the presently studied detection problem. 
The mAP is the average of all APs for a particular class which can be expressed as, 
\begin{equation}
	mAP= \frac{1}{N} \sum^N  AP
	\label{E-10}
\end{equation} 

\section{Results \& Discussion}
\label{sec:5}

\begin{figure}
	\noindent
	\centering
	\includegraphics[width=0.8\linewidth]{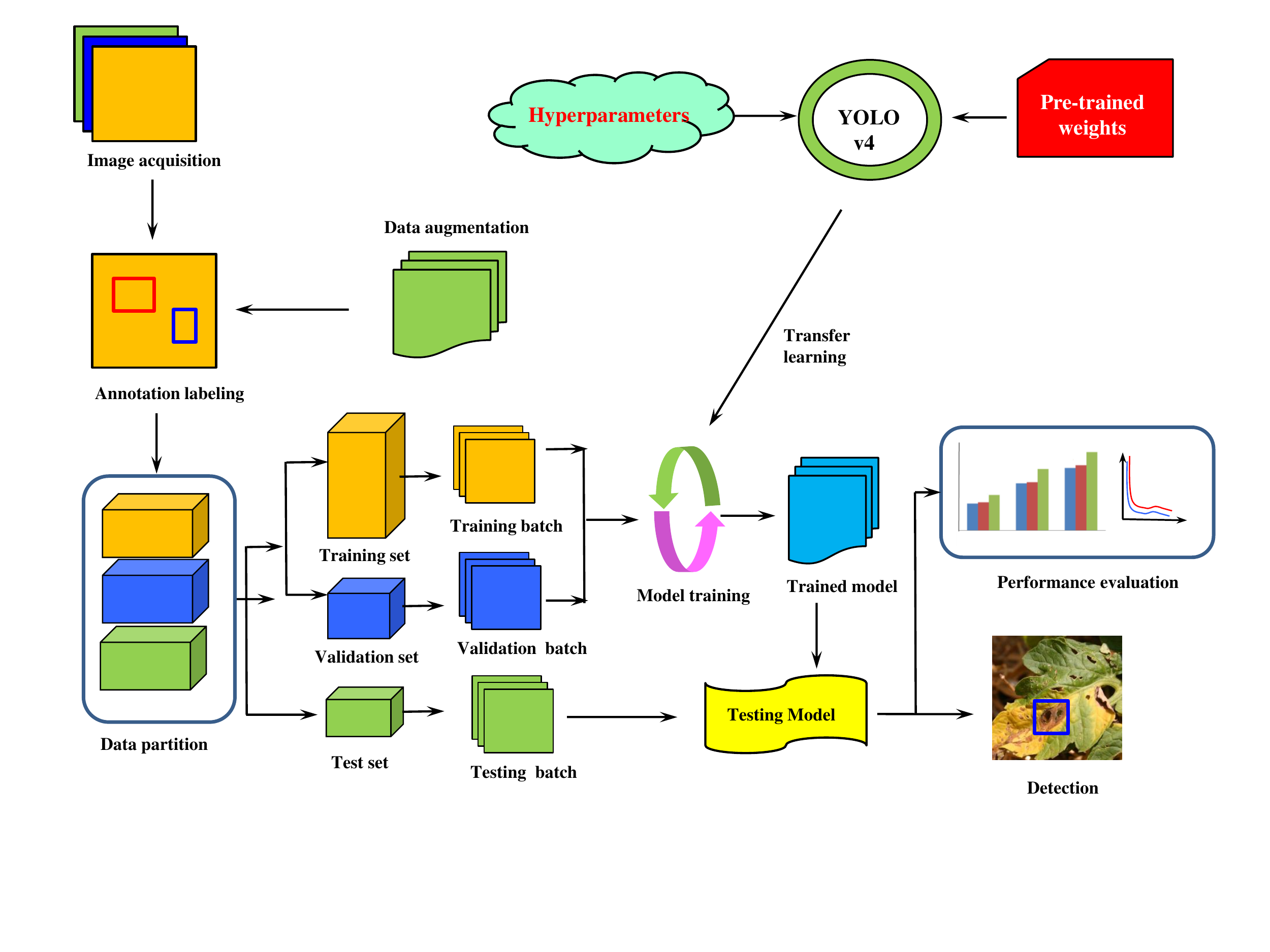}
	\caption{\label{Fig-7}Flowchart of the  overall workflow methodology for the proposed detection model. }
\end{figure}

The state-of-the-art YOLOv4 algorithm is utilized to develop an accurate real‑time high-performance image detection model on a single GPU. 
In this section the proposed model is tested on the tomato plant disease detection problem. 
A flowchart of the detection model workflow methodology is shown in Fig. \ref{Fig-7}. 
At first, 300 images from each of the four different tomato plant diseases (i.e., early blight, late blight, Septoria leaf spot, and leaf mold) are collected from the publicly available popular Kaggle PlantVillage Dataset \cite{plantvillage} to construct a dataset consisting of 1200  images. 
To improve robustness and avoid over-fitting during training, the dataset is expanded ten folds using different image augmentation procedures to obtain the custom dataset of 12,000 images. 
For image annotation of the target classes in the custom dataset, a Python-based open-source script, LabelImg \cite{LabelImg} is used which saves the annotations as XML files and organizes them into PASCAL VOC format. 
Each XML file contains information of the target class and corresponding bounding box coordinates for the training image dataset. 
From the custom dataset, a total of 8,400, 1,800, and 1,800 are randomly chosen as training, validation, and test images, respectively. 
The model is trained through a transfer learning method initialized from a pre-trained weights-file \cite{AlexeyAB}. 
Training and testing are performed on the local system utilizing a single NVIDIA GeForce RTX 2080 GPU. 
Computing resources and DNN environment specifications used for these purposes are listed in Table \ref{T-1}. 

\begin{table}
	\centering
	\caption{Local computing resources and DNN environments}
	\begin{tabular}{c c  }
		\\[-0.5em]
		\hline
		\\[-0.8em] 
		Testing Environment  $\quad $ & Configuration Parameter  
		\\[-0.0em]
		\hline
		\\[-0.5em]
		OS & Windows 10 Pro 64
		\\
		\\[-0.5em]
		CPU  & Intel Core i5   
		\\
		\\[-0.5em]
		RAM   & 8 GB DDR4 
		\\
		\\[-0.5em]
		GPU  & NVIDIA GeForce RTX 2080 
		\\
		\\[-0.5em]
		GPU acceleration env.   & CUDA 10.2.89
		\\
		\\[-0.5em]
		GPU accelerated DNN lib.  &  cuDNN 10.2 v7.6.5
		\\
		\\[-0.5em]
		Python distrib.   & Anaconda 3 2019.10
		\\[-0.5em]
		\\
		ML lib.   & Keras, Tensorflow, PyTorch
		\\
		\\[-0.5em]
		Int.  development env.   & Visual Studio comm. v15.9 (2017)
		\\
		\\[-0.5em]
		Comp. Vision lib.     & OpenCV 4.5.1-vc14   
		\\
		\\[-0.5em]
		\hline
	\end{tabular}
	\label{T-1}
\end{table} 
Configuration parameters, such as, number of channels, momentum value, decay regularization, etc. are same as in the original YOLOV4 model. 
Primary hyperparameters used for training are summarized in Table \ref{T-2}. 
\begin{table}
	\centering
	\caption{Initial  configuration parameters of the proposed detection model}
	\begin{tabular}{c c c c c}
		\\[-0.5em]
		\hline
		\\[-0.8em] 
		Input size of image  $\quad $ & Batch   $\quad $ & Subdivision  $\quad $  & Channels $\quad $ & Momentum  $\quad $ 
		\\[-0.0em]
		\hline
		\\[-0.5em]
		$416\times416$ & 16 & 4 & 3 & 0.9
		\\[-0.5em]
		\\
		\hline
		\\[-0.8em] 
		Class coeff.   $\quad $ & Obj coeff.  $\quad $ & Shear  $\quad $  & Mosaic $\quad $ & Mix-up  $\quad $ 
		\\[-0.0em]
		\hline
		\\[-0.5em]
		0.5 & 1 & 0 & 1 & 0
		\\[-0.5em]
		\\
		\hline
		\\[-0.8em] 
		Initial learning rate   $\quad $ & Decay   $\quad $ & Classes  $\quad $  & Filters $\quad $ & Training steps  $\quad $ 
		\\[-0.0em]
		\hline
		\\[-0.5em]
		0.001 & 0.005 & 4 & 27 & 85,000
		\\[-0.5em]
		\\
		\hline
	\end{tabular}
	\label{T-2}
\end{table}


\subsection{Detection accuracy of the proposed model}
\label{sec:5a}

Several issues can hinder the performance of the YOLOv4 model in detecting diseases in tomato plants, such as, densely populated fine-grain diseases, irregular geometric morphology of infected areas, coexistence of multi-scale disease spots, similarity of color textures of infected areas and leaves, and varying lightening conditions resulting in low detection accuracy, a high number of missed detections, and false object prediction. 
In order to overcome such issues, modifications of the detection model proposed in Section 3 yields improved feature map in complex zones increasing the accuracy of the bounding boxes. 
\begin{table}
	\centering
	\caption{Comparison of  network performance considering different activation functions in YOLO backbone-neck combinations  with integrated SPP plugin for anchors of $416\times416$ input resolution    tested in NVIDIA GeForce RTX 2080 GPU. All APs are in $\%$. }
	\begin{tabular}{c c c c c c c c c c }
		\\[-0.5em]
		\hline
		\\[-0.8em] 
		\vtop{\hbox{\strut Backbone +}\hbox{\strut Activation}}& \vtop{\hbox{\strut Neck+}\hbox{\strut Activation}}  & $AP$  & $AP_{50}$ & $AP_{75}$ & $AP_{S}$ & $AP_{M}$ & $AP_{L}$ & FPS
		\\[-0.0em]
		\hline
		\\[-0.5em]
		CSPDarknet53+L-ReLU & PANet+L-ReLU & 60.3 & 87.6 & 74.9& 43.5  & 67.3  & 66.6 & 67.9 
		\\
		\\[-0.5em]
		CSPDarknet53+Mish & PANet+L-ReLU  & 64.2 & 92.8 & 78.2  & 48.3  & 63.7  & 70.5 & 63.6 
		\\
		\\[-0.5em]
		CSPDarknet53+H-swish & PANet+H-swish  & 65.8 & 93.1 & 77.5  & 49.9  & 61.2  & 65.9 & 78.2 
		\\
		\\[-0.5em]
		D-CSPDarknet53+L-ReLU & PANet+L-ReLU & 67.3 & 94.5 & 76.7& 46.5  & 69.8  & 64.6 & 61.2 
		\\
		\\[-0.5em]
		D-CSPDarknet53+Mish & PANet+L-ReLU & 70.1 & 94.9 & 78.8& 51.5  & 71.5  & 66.7 & 59.7 
		\\
		\\[-0.5em]
		D-CSPDarknet53+Mish & PANet+Mish & 71.5 & 95.7 & 79.2  & 52.8 & 69.9  & 68.2 & 58.9 
		\\
		\\[-0.5em]
		D-CSPDarknet53+H-swish & PANet+Mish & 74.8 & 96.1 & 82.2  & 71.8 & 76.9  & 51.7 & 67.9 
		\\
		\\[-0.5em]
		\bf{D-CSPDarknet53+H-swish} & \bf{PANet+H-swish} & \bf{79.6}  & \bf{96.3} & \bf{91.6}  & \bf{73.6} & \bf{82.9}  & \bf{89.5} & \bf{70.2} 
		\\
		\\[-0.5em]
		\hline
	\end{tabular}
	\label{T-3}
\end{table}

\subsubsection{Influence of different activation functions}
\label{sec:5ai}
\begin{figure}
	\noindent
	\centering
	\includegraphics[width=1.15\linewidth]{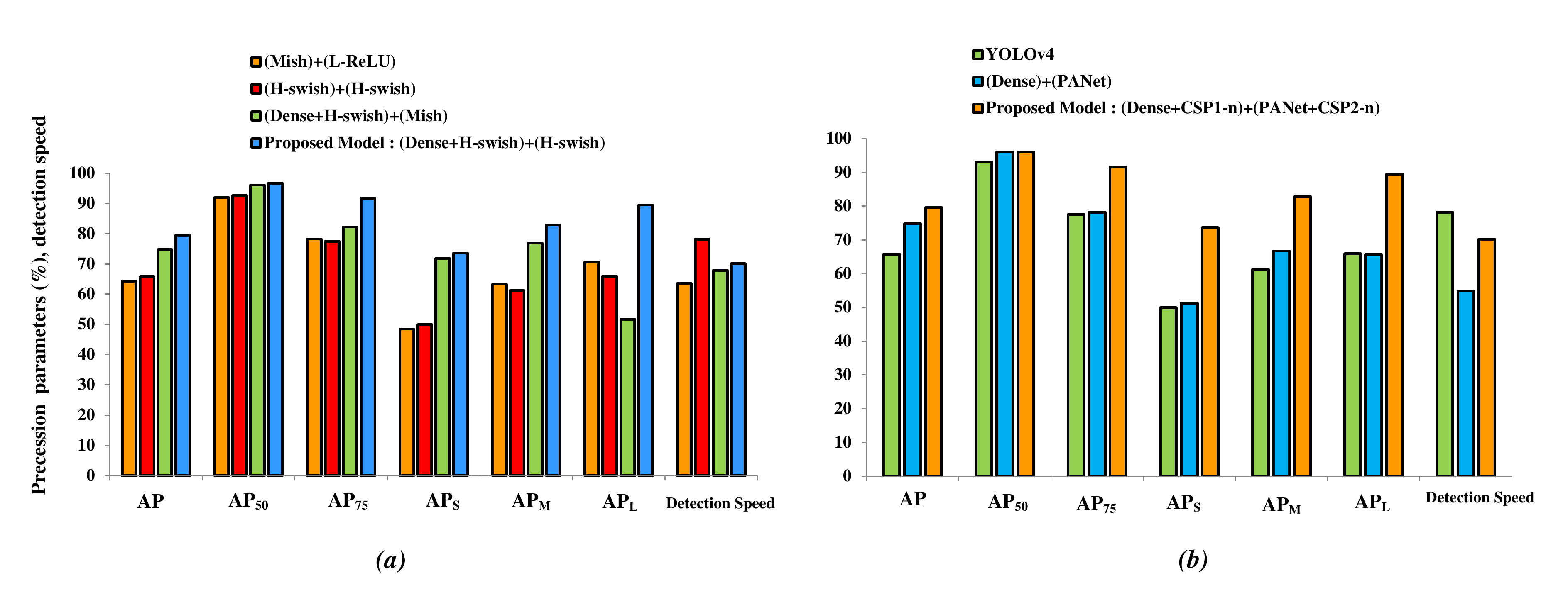}
	\caption{\label{Fig-8} 
		Comparison barchart of different precision parameters and detection speed (in FPS) for (a) different activation functions; (b) different combinations  of CSP1-$n$ in backbone and CSP2-$n$ in neck of the detection model.}
\end{figure}

To obtain the best detection model, the accuracy and detection speed for different combinations of the backbone structure and the neck configuration are tested. 
Three activation functions ( Leaky-ReLU, Mish, H-swish) are chosen for the backbone and neck parts of the model and all possible combinations are considered in comparing different precision parameters (i.e., AP, AP$_{50}$, AP$_{75}$, AP$_{S}$, AP$_{M}$, and AP$_{L}$) and corresponding detection speed. 
Results are presented in Table .\ref{T-3}. 
The target confidence and $IoU$ threshold were prescribed as  0.3 and 0.5, respectively. 
The Leaky-ReLU as the primary activation function in both backbone and neck parts of the detection model provides the least accurate model among all the variants. 
Compared to Leaky-ReLU, implementation of the Mish activation function increases the AP values at the expense of detection speed. 
The accuracy is further improved upon implementation of dense blocks in the backbone for both Leaky-ReLU and Mish activation functions. 
The dense block provides significant improvement in detection accuracy of the model for both cases while the detection speed is slightly compromised. 
However, together with dense blocks, implementation of the H-swish as the primary activation function in both backbone and neck provides the best results in terms of detection accuracy and speed as shown in Fig. \ref{Fig-8}-(a). 
Upon using the H-swish function instead of the Leaky-ReLU as the activation, a significant gain in accuracy is obtained on the custom dataset as AP increases from  60.3$\%$ to 79.6$\%$,  AP$_{50}$ increases from  87.6$\%$ to 96.3$\%$, AP$_{75}$ increases from  74.9$\%$ to 91.6$\%$, AP$_{S}$ increases from  43.5$\%$ to 73.6$\%$ and AP$_{M}$ increases from  67.3$\%$ to 82.9$\%$; AP$_{L}$ increases from  66.6$\%$ to 89.5$\%$ as shown in Table \ref{T-3}. 
Improvement of different AP values also indicates a stronger nonlinear feature learning capability with the H-swish function compared to Leaky-ReLU and Mish functions. 
Moreover, with the introduction of dense blocks, integration of the SPP module, and modification of the PANet in the proposed model, the fine-grain feature transfer and reuse improve. 
This leads to improved detection accuracy for small and medium object detection (i.e., AP$_{S}$ and AP$_{M}$) on the feature data set. 
Regarding detection speed, using the Mish activation for both backbone and neck adversely affects the detection speed compared to the Leaky-ReLU. 
However, H-swish activation provides the fastest detection as the detection speed increases by 14.7$\%$ compared to the Leaky-ReLU activation. 
Evidently, the H-swish activation function provides superior performance (both detection accuracy and speed) on the disease dataset. 
\begin{table}
	\centering
	\caption{Influence of CSP1-$n$ and CSP2-$n$ blocks  on the   network performance for the combinations of YOLO backbone-neck add-in   considering H-swish as primary  activation function  with integrated SPP plugin   for anchors of $416\times416$. 
	}
	\begin{tabular}{c c c c c c c c c c }
		\\[-0.5em]
		\hline
		\\[-0.8em] 
		\vtop{\hbox{\strut Backbone}\hbox{\strut+ add-in}}& \vtop{\hbox{\strut Neck}\hbox{\strut +add-in}}  & $AP$  & $AP_{50}$ & $AP_{75}$ & $AP_{S}$ & $AP_{M}$ & $AP_{L}$ & FPS
		\\[-0.0em]
		\hline
		\\[-0.5em]
		CSPDarknet53 & PANet & 65.8 & 93.1 & 77.5  & 49.9  & 61.2  & 65.9 & 78.2 
		\\
		\\[-0.5em]
		D-CSPDarknet53 & PANet  & 74.8 & 96.1 & 78.2  & 51.3  & 66.7  & 65.7 & 54.9 
		\\
		\\[-0.5em]
		D-CSPDarknet53+CSP1-$n$ & PANet & 77.5 & 96.2 & 81.5  & 63.9  & 68.2  & 70.9 & 66.2 
		\\
		\\[-0.5em]
		CSPDarknet53 & PANet+CSP2-$n$ & 73.8 & 95.6 & 71.2  & 54.8 & 67.9  & 74.7 & 78.9 
		\\
		\\[-0.5em]
		D-CSPDarknet53+CSP1-$n$ & PANet+CSP2-$n$ & \bf{79.6}  & \bf{96.3} & \bf{91.6}  & \bf{73.6} & \bf{82.9}  & \bf{89.5} & \bf{70.2} 
		\\
		\\[-0.5em]
		\hline
	\end{tabular}
	\label{T-4}
\end{table}

\subsubsection{Influence of residual blocks}
\label{sec:5aii}

The influence of inclusions of the CSP1-$n$ in the backbone and the CSP2-$n$ in the modified  PANet on the network performance is reflected in the accuracy measures reported in Table.\ref{T-4}. 
For comparison, five different combinations of backbone and neck are considered with or without the additional residual blocks. 
Without the use of the CSP1-$n$ and CSP2-$n$, the detection model provides the worst results in terms of accuracy among all variants. 
Introducing dense and CSP1-$n$ blocks in backbone increases all accuracy parameters, in particular, the AP$_{50}$ and AP$_{S}$ as shown in Fig. \ref{Fig-9}-(b). 
Such a combination though reduces the detection speed compared to the other variants. 
However, implementation of the CSP2-$n$ in addition to the CSP1-$n$ and dense blocks, both detection accuracy and speed increase; AP, AP$_{75}$, AP$_{S}$, AP$_{M}$ and AP$_{L}$ increase by 2.1$\%$, 14.8$\%$, 9.7$\%$, 14.7$\%$ and 18.6$\%$, respectively. 
Also, 6.3$\%$ increase in detection speed is obtained as shown in Fig. \ref{Fig-9}-(b). 
Evidently, use of the CSP1-$n$ block in the backbone enhances feature extraction and use of the CSP2-$n$ block in the PANet enhances the learning capability of semantic features increasing AP values and detection speed. 
The comparisons demonstrate the effectiveness of the CSP1-$n$ and CSP2-$n$ modules in the proposed model. 

\subsection{Comparison with existing state-of-the-art models}
\label{sec:5b}

Detection performance of the proposed model is evaluated by comparing different metrics for accuracy ($P$, $R$, $F1-$score, $mAP$ at $IoU\geq 0.5$, and detection speed as in section 4.2) with existing state-of-the-art models for object detection \cite{Zhao-IEEE-2019}. 
YOLOv3, YOLOv4, and the proposed model (improved YOLOv4) along with five additional detection models: Faster RCNN \cite{Ren_et_al-2015}, RetinaNet \cite{Lin1-IEEE-2017}, 
single shot multi-box detector (SSD) \cite{Liu-ECCV-2016}, Cascade R-CNN \cite{Cai-IEEE-2019}, and the Mask R-CNN \cite{He_et_al-2017} are trained and tested on the custom disease dataset in an open-source OpenMMLab object detection toolbox (MMDetection) \cite{MMDetection} based on PyTorch. 
The hyperparameters were kept as consistent as possible while comparing other models with our proposed model. 
The performance metrics for the 8 models considered herein are tabulated in Table \ref{T-5}. 
\begin{table}
	\centering
	\caption{Comparison of precision, recall, F1-score, mAP, and detection speed (in FPS) between proposed model (improved YOLOv4) and other state-of-the-art models: Faster R-CNN, RetinaNet,  SSD, Mask R-CNN, Cascade R-CNN, YOLOv3, and YOLOv4 tested in NVIDIA GeForce RTX 2080 GPU. Bold highlights the best result obtained from corresponding model performances.}
	\begin{tabular}{c c c c c c c c }
		\\[-0.5em]
		\hline
		\\[-0.8em] 
		Model & P ($\%$) & R ($\%$)  &F1-score ($\%$)  & mAP ($\%$)  & Dect. time (ms) & FPS
		\\[-0.0em]
		\hline
		\\[-0.5em]
		Faster R-CNN & 26.27 & 42.39 & 55.73  & 59.17  & 44.42 & 22.51   
		\\
		\\[-0.5em]
		RetinaNet & 43.28 & 51.77 & 60.42  & 63.78  & 30.52& 32.82 
		\\
		\\[-0.5em]
		SSD & 65.45 & 76.34 & 80.97  & 82.52  & 25.71& 38.39 
		\\
		\\[-0.5em]
		Mask R-CNN & 63.52 & 75.35 & 80.23  & 83.61 &55.82 & 19.75 
		\\
		\\[-0.5em]
		Cascade R-CNN & 70.42 & 79.25 & 84.13  & 86.61 &35.32&  28.13 
		\\
		\\[-0.5em]
		YOLOv3 & 73.98 & 88.01 & 80.38  & 89.19 &16.52& 60.57
		\\
		\\[-0.5em]
		YOLOv4 & 81.39 & 92.14 & 86.44  & 92.84 &15.72& 63.61
		\\
		\\[-0.5em]
		\bf{Proposed model} & \bf{90.33} & \bf{97.20} & \bf{93.64}  & \bf{96.29} &\bf{14.29}& \bf{70.19 }
		\\
		\\[-0.5em]
		\hline
	\end{tabular}
	\label{T-5}
\end{table}
\begin{figure}
	\noindent
	\centering
	\includegraphics[width=1.1\linewidth]{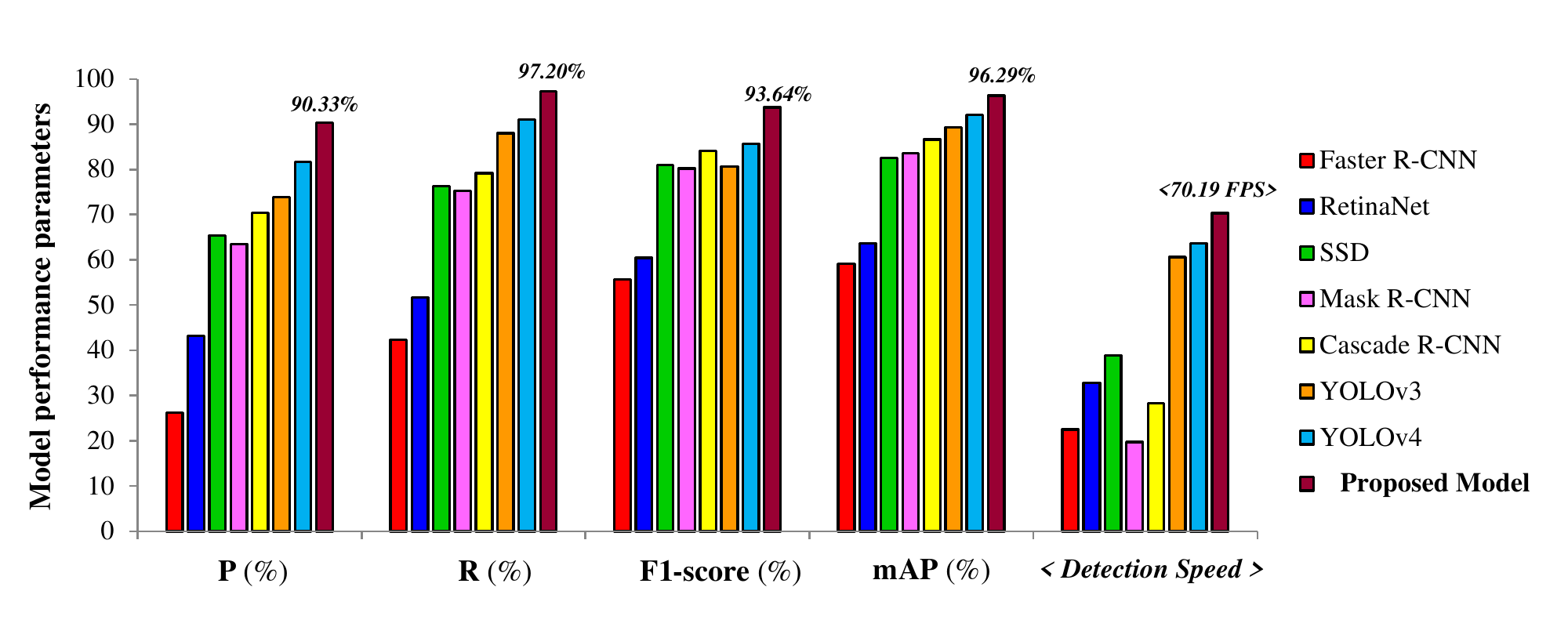}
	\caption{\label{Fig-9} Comparison bar chart of precision, recall, F1-score, mAP, and detection speed (in FPS) between proposed model (improved YOLOv4) and other state-of-the-art models: Faster R-CNN, RetinaNet,  SSD, Mask R-CNN, Cascade R-CNN, YOLOv3, and YOLOv4.}
\end{figure}
The bar chart plotted in Fig. \ref{Fig-9} shows that the performance of the Faster R-CNN and the RetinaNet is inferior to other models on the custom disease dataset for all metrics with precision values of $26.27\%$ and $43.28\%$, respectively. 
The SSD and the Mask R-CNN demonstrate better performance compared to the RetinaNet with  $20.55\%$ and  $19.81\%$ increase in $F1-$score and $18.74\%$ and $19.83\%$ increase in $mAP$, respectively. 
Additionally, the detection speed recorded for Faster R-CNN, RetinaNet, SSD, Mask R-CNN,
and Cascade R-CNN are relatively low indicating limitation of these models for real-time on-field object detection tasks at high resolution. 
Although the Cascade  R-CNN shows some improvement in detection accuracy compared to the previous four models, performance parameters obtained from these models are significantly lower compared to the versions of the original YOLO (i.e.,  YOLOv3 and YOLOv4) and the proposed detection model for the custom disease dataset used herein. 
For example, YOLOv4 yields $10.97\%$ and  $12.89\%$ increase in precision and recall values compared to the Cascade R-CNN. 
However, our proposed model is superior to YOLOv4 with $8.94\%$, $5.06\%$, $7.20\%$, and  $3.45\%$ increase in precision, recall, $F1-$score, and $mAP$, respectively as shown in Fig.\ref{Fig-9}. 
Moreover, the proposed model is faster at detection at 70.19 FPS  which is $10.34\%$ higher than the original YOLOv4 model. 
Region based detection model such as Faster RCNN, RetinaNet, SSD perform classification and bounding box regression with its two step architecture. Whereas, YOLO makes classification and bounding box regression at the same time making the detection model significantly faster.
Summarizing, the proposed detection model outperforms other state-of-the-art models in both detection accuracy and speed making it a promising model for high-performance real-time disease detection for precision agriculture and automation. 
\begin{table}
	\centering
	\caption{Comparison of IoU, F1 Score, final loss, detection speed  and average detection time between YOLOv3, YOLOv4, and proposed model.}
	\begin{tabular}{c c c c  c c c }
		\\[-0.5em]
		\hline
		\\[-0.8em] 
		Detection model  &IoU  & F1-score  & Validation loss   & \vtop{\hbox{\strut Detection }\hbox{time (ms) }} & \vtop{\hbox{\strut Detection }\hbox{speed (FPS) }} & Training time
		\\[-0.0em]
		\hline
		\\[-0.5em]
		YOLOv3 & 0.767 & 0.803 & 13.31 & 16.52 & 60.57 & 7.35h
		\\
		\\[-0.5em]
		YOLOv4 & 0.872 & 0.864 &  8.92 & 15.72 & 63.61 & 6.14h
		\\
		\\[-0.5em]
		Proposed model & 0.935 & 0.936  & 2.29 &14.29  & 70.19 & 3.98h
		\\
		\\[-0.5em]
		\hline
	\end{tabular}
	\label{T-6}
\end{table}
\begin{figure}
	\noindent
	\centering
	\includegraphics[width=1.1\linewidth]{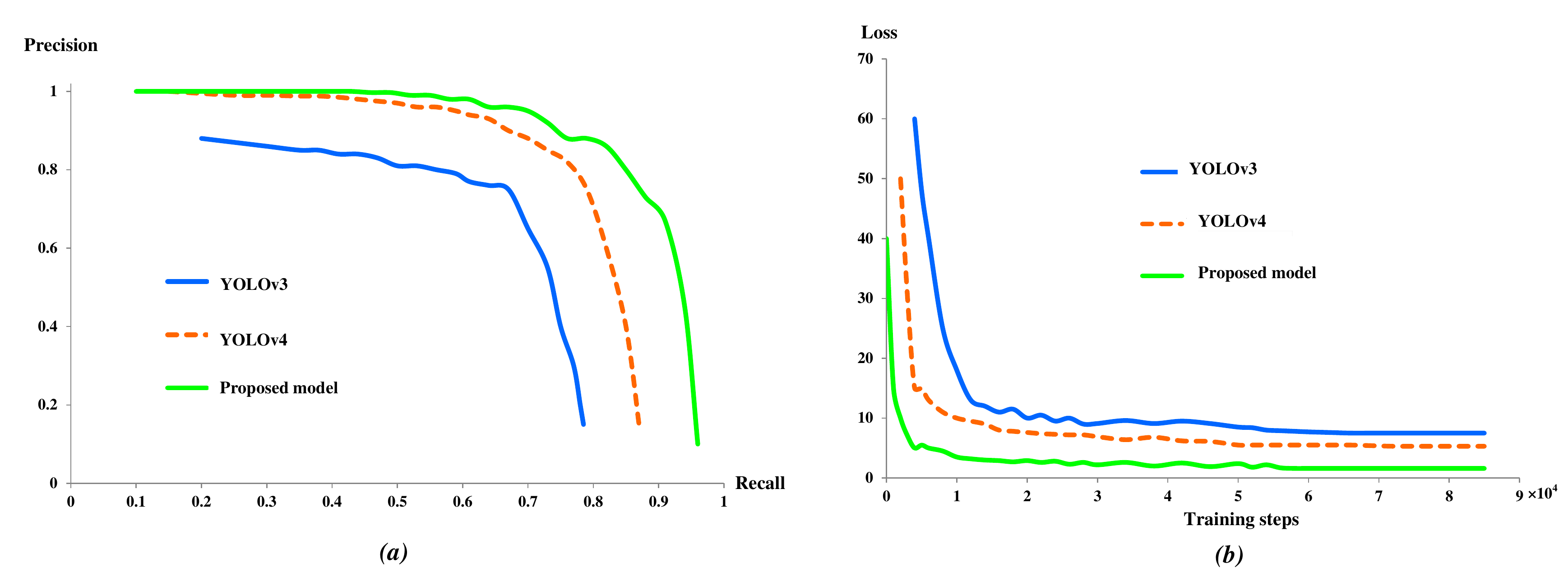}
	\caption{\label{Fig-10} Comparison of (a) P-R curves during testing ; (b) training loss curves  between  YOLOv3, YOLOv4, and proposed detection model.}
\end{figure}

\subsection{Overall performance of the proposed detection model}
\label{sec:5c}

From the previous section, we identify the YOLOv3 and YOLOv4 models as comparable to the proposed model in detection accuracy and speed. 
These three models are extensively tested and compared in this section on the disease dataset. 
The values of $IoU$, $F1$ scores, final loss, and average detection time for these three models are compared in Table \ref{T-6} for the test images. 
YOLOv3 yields the lowest $IoU$ value of 0.767 between the three models, whereas, the $IoU$ obtained for the proposed model is the maximum at 0.935, which is 10.4\% higher than the original YOLOv4. 
The $IoU$ values suggest that the proposed detection model is the most accurate at detecting bounding boxes. 
The original YOLOv4 with an $F1-$score of 0.861 is more efficient in disease detection than the YOLOv3 ($F1-$score of 0.803). 
However, the proposed model also yields the highest $F1$ score of 0.936 which is an improvement of $7.20\%$ from the original YOLOv4. 
The proposed model also provides the best precision and recall performance compared to the other two models (see Table \ref{T-7}). 
Furthermore, comparison of the average detection time reveals that the proposed model has the lowest detection time of 14.29 $ms$ resulting in the fastest detection speed of 70.19  FPS as listed in Table \ref{T-6}. 
Besides, the proposed model is comparatively easy to train (training time s 3.98 $h$ which is the lowest). 
Therefore, it is evident that the proposed object detection model outperforms the YOLOv3 and YOLOv4 in all performance measures. 
Comparison of the precision-recall ($PR$) curves for these three models are shown in Fig.~\ref{Fig-10}-(a). 
Comparing the characteristics of the $P-R$ curves, the precision for the proposed model is higher than YOLOv3 and YOLOv4 models for any recall. 
The area under the $P-R$ curve (i.e., $AP$ value) is the maximum for the proposed model  (see Table \ref{T-5} for comparison of $mAP$ at $IoU \geq 0.5$) which suggests that the proposed model exhibited better accuracy and precision at detection compared to the YOLOv3 and YOLOv4. 
\begin{figure}
	\noindent
	\centering
	\includegraphics[width=1.1\linewidth]{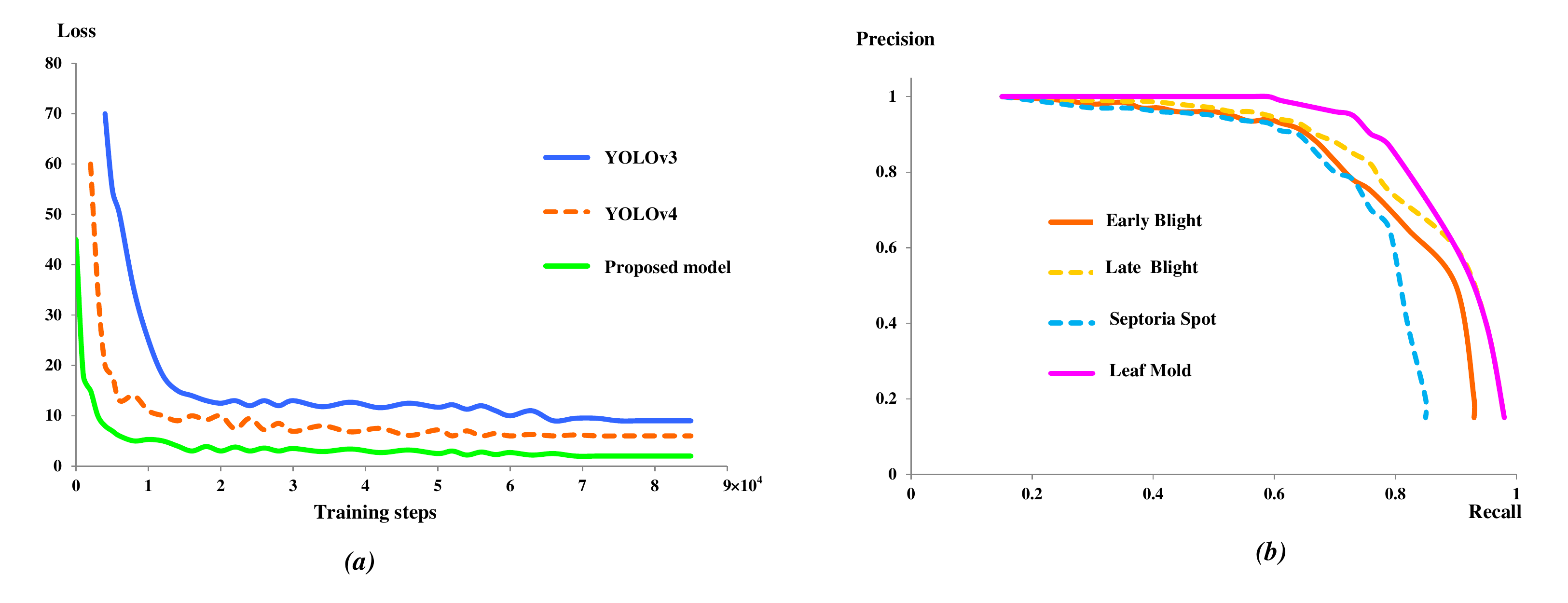}
	\caption{\label{Fig-11} (a) Comparison of validation loss curves between  YOLOv3, YOLOv4, and improved YOLOv4 detection models; (b) P-R curves for different disease class from the proposed model.  }
\end{figure}

Figure \ref{Fig-10}-(b) compares the training loss plotted for the YOLOv3, YOLOv4, and modified YOLOv4 models. 
In the initial phase, the loss curves corresponding to the YOLOv3 and YOLOv4 begin to reduce significantly after approximately 16,000 and 11,000 training steps. 
For the proposed model, loss reduction is much faster up to at least 5,000 training steps. 
After exhibiting several cycles of fluctuation, training loss in the proposed model tends to saturate after approximately 55,000 training steps. 
The final training loss for the proposed model is 2.98 which is approximately $34.7\%$ lower than the original YOLOv4. 
Usually, a lower training loss yields a more accurate model. 
Similarly characteristics are obtained for the validation loss curve plotted in Fig. \ref{Fig-11}-(a); validation loss for the proposed model decreases rapidly up to 3,000  training steps. 
The loss curve saturates as the model gradually converges after about 65,000 steps. 
After 85,000 training steps, both training and validation losses are observed to saturate. 
The obtained model is therefore assumed to be optimal. 
The final validation loss value for the proposed model is 2.29 compared to 13.31 and 8.92 for the YOLOv3 and YOLOv4, respectively as shown in Table \ref{T-6}. 
The proposed model is evidently easier to train, converging faster to a more accurate model. 

How do the model perform in detecting individual classes? 
$TP$, $FP$, and $FN$ for each class and corresponding precision, recall and $F1-$score are tabulated in Table \ref{T-7}. 
For comparison, $P-R$ curves for different disease classes for the proposed model are plotted in Fig. \ref{Fig-11}-(b) which indicate that the model detects leaf mold better (as the precision-recall values are higher) than the other classes. 
The proposed model has relatively higher precision for late blight ($91.76\%$) and Septoria ($94.89\%$). 
The proposed model yields $90.33\%$ precision, $97.20\%$ recall, and $93.64\%$ $F1-$score. 
In comparison to other models, the model maximizes the $TP$ value minimizing $FP$ and $FN$ for all classes compared to YOLOv3 and YOLOv4. 
For example, $TP$ increases from 9575 to 10780; $FP$ and $FN$ reduce from 2188 to 1153 and 816 to 310, respectively, in Table \ref{T-7}. 
This results in  $8.94\%$, $5.06\%$, $7.20\%$, and  $3.45\%$ increase in precision, recall, $F1-$score, and $mAP$, respectively, compared to the original YOLOv4. 

\begin{table}[]
	\centering
	\caption{Comparison of detection results between YOLOv3, YOLOv4, and proposed  model on the test dataset. }
	\label{T-7}
	\begin{tabular}{c c c c c  c c c c}
		\\[-0.5em]
		\hline
		\\[-0.8em] 
		Model  & Class  & Objects & TP  &FP& FN & P ($\%$) & R ($\%$) & F1-score  \\ \hline
		{\vtop{\hbox{\strut YOLOv3}\hbox{\strut\,\, }}} 
		&\bf{All}&\bf{11952} &\bf{8508} &\bf{2991} &\bf{1160}&\bf{73.98} &\bf{88.01}&\bf{80.38}
		\\ \cline{2-9} 
		&Early Blight&5085 &3658 &1412 &510 &72.14 &87.76&79.19
		\\ \cline{2-9} 
		&Late Blight&2304 &1701 &511 &209 &76.92 &89.05&82.54
		\\ \cline{2-9} 
		&Septoria&2799 &1789 &818 &343 &68.62 &83.91&75.50
		\\ \cline{2-9} 
		&Leaf Mold &1764 &1361 &251 &98 &84.42 &93.28&88.63
		\\ \hline
		{\vtop{\hbox{\strut YOLOv4}\hbox{\strut\,\, }}} 
		&\bf{All}&\bf{11952} &\bf{9575} &\bf{2188} &\bf{816} &\bf{81.39}&\bf{92.14}&\bf{86.44}
		\\ \cline{2-9} 
		&Early Blight&5085 &4058 &1002 &375 &80.19 &91.54&85.49
		\\ \cline{2-9} 
		&Late Blight&2304 &2007 &396 &141 &83.52&93.43&88.20
		\\ \cline{2-9} 
		&Septoria&2799 &2041 &593 &255 &77.48 &88.89&82.79
		\\ \cline{2-9} 
		&Leaf Mold &1764 &1469 &197 &45 &88.17 &97.02& 92.38
		\\ \hline
		{\vtop{\hbox{\strut Proposed }\hbox{\strut\,\,Model }}} 
		&\bf{All}&\bf{11952} &\bf{10780} &\bf{1153} &\bf{310} &\bf{90.33} &\bf{97.20}& \bf{93.64}
		\\ \cline{2-9} 
		&Early Blight&5085 &4578 &476 &157 &90.58 &96.68&93.53
		\\ \cline{2-9} 
		&Late Blight&2304 &2196 &197 &41 &91.76 &98.16&94.85
		\\ \cline{2-9} 
		&Septoria&2799 &2334 &390 &96 &85.68 &96.04&90.57
		\\ \cline{2-9} 
		&Leaf Mold &1764 &1672 &91 &16 &94.89 &99.05 &96.92
		\\ \hline
	\end{tabular}
\end{table}

\begin{table}[]
	\centering
	\caption{Comparison of detection performance and detection speed between YOLOv4 and the proposed model  for  GIoU, DIoU, and CIoU loss functions. }
	\label{T-8}
	\begin{tabular}{c c c c c  c c c}
		\\[-0.5em]
		\hline
		\\[-0.8em] 
		Model  & Class  & P ($\%$) & R ($\%$)  &F1 & mAP ($\%$) & Dect. time (ms) & FPS  \\ \hline
		{\vtop{\hbox{\strut YOLOv4}\hbox{\strut\,\, (GIoU)}}} 
		&All&78.71 &89.56 &83.79 &90.34 
		\\ \cline{2-6} 
		&Early Blight&74.90 &87.82 &80.84 &90.68 
		\\ \cline{2-6} 
		&Late Blight&78.45 &90.28 &83.95 &89.93 &14.34 &69.7
		\\ \cline{2-6} 
		&Septoria &77.53 &87.30 &82.12&88.98 
		\\ \cline{2-6} 
		&Leaf Mold &83.98 &92.87 &88.20 &91.79
		\\ \hline
		{\vtop{\hbox{\strut Proposed  model}\hbox{\strut\,\,\,\,\,\,\,\,\,(GIoU)}}} 
		&All&88.55 &93.39 &90.91 &94.16
		\\ \cline{2-6} 
		&Early Blight&87.35 &93.12 &90.14 &94.79
		\\ \cline{2-6} 
		&Late Blight&91.47 &93.27 &92.36&97.01 &13.26 &75.43
		\\ \cline{2-6} 
		&Septoria &82.73 &91.89 &87.07 &88.97
		\\ \cline{2-6} 
		&Leaf Mold &92.67 &95.31 &93.97 &95.89 
		\\ \hline
		{\vtop{\hbox{\strut YOLOv4}\hbox{\strut\,\, (DIoU)}}} 
		&All& 80.96 &91.82 &86.05 &92.41 
		\\ \cline{2-6} 
		&Early Blight&78.54 &90.45 &84.07 &92.76
		\\ \cline{2-6} 
		&Late Blight&81.24 &91.07 &85.87 &92.98 &14.59 &68.52
		\\ \cline{2-6} 
		&Septoria &78.21 &89.89 &83.64 &91.03 
		\\ \cline{2-6} 
		&Leaf Mold &85.87 &95.87 &90.59 &92.87
		\\ \hline
		{\vtop{\hbox{\strut Proposed model}\hbox{\strut\,\,\,\,\,\,\,\,\,(DIoU)}}} 
		&All&89.43 &94.83 &92.05 &94.76 
		\\ \cline{2-6} 
		&Early Blight&90.12 &94.53 &92.27 &95.32
		\\ \cline{2-6} 
		&Late Blight&90.92 &92.86 &91.88 &96.56 &13.66 &73.21
		\\ \cline{2-6} 
		&Septoria &84.13 &94.37 &88.95 &90.87 
		\\ \cline{2-6} 
		&Leaf Mold &92.56 &97.59&95.01 &96.32 
		\\ \hline
		{\vtop{\hbox{\strut YOLOv4}\hbox{\strut\,\, (CIoU)}}} 
		&All&81.39 &92.14&86.44 &92.84 
		\\ \cline{2-6} 
		&Early Blight&80.19 &91.54&85.49 &93.05
		\\ \cline{2-6} 
		&Late Blight&83.52 &93.43 &88.20 &94.14 &15.72 &63.61
		\\ \cline{2-6} 
		&Septoria &77.48 &88.89 &82.79 &87.41 
		\\ \cline{2-6} 
		&Leaf Mold &88.17 &97.02 &92.38 &96.76 
		\\ \hline
		{\vtop{\hbox{\strut Proposed model}\hbox{\strut\,\,\,\,\,\,\,\,\, (CIoU)}}} 
		&All&90.33 &97.20 &93.64 & 96.29 
		\\ \cline{2-6} 
		&Early Blight&90.58 &96.68 &93.53 &96.01 
		\\ \cline{2-6} 
		&Late Blight&91.76 &98.16 &94.85 &97.98 &14.29 &70.19
		\\ \cline{2-6} 
		&Septoria &85.68 &96.04 &90.57 &92.45
		\\ \cline{2-6} 
		&Leaf Mold &94.89 &99.05 &96.92 &98.72 
		\\ \hline
		
	\end{tabular}
\end{table}

Influence of using different loss functions, namely, $GIoU$, $DIoU$, and $CIoU$ losses on detection performance and speed is compared for YOLOv4 and the proposed model in Table \ref{T-8}. 
Choosing appropriate loss function can improve the detection accuracy at the same time maintaining high detection speed during disease detection. 
Among the three loss functions, employing the $CIoU$ loss results in obtaining the best network. 
For example, precision, recall, $F1$, and $mAP$ increase by $2.78\%$, $3.81\%$, $2.73\%$, and  $1.5\%$, respectively compared to the model obtained using the $GIoU$ loss. 
However, using the $CIoU$ slightly compromises the detection speed while the detection time increases by only 1.03 $ms$ compared to the $GIoU$ loss. 
Overall, a slight improvement in performance is obtained for all disease classes when the $CIoU$ loss function is used for both YOLOv4 and the proposed model. 
Although the use of an appropriate loss function has some influence, it is relatively less compared to the use of appropriate activation function, addition of the residual module and dense block. 
Nevertheless, appropriate selection of both activation and loss functions lead to maximizing overall network accuracy and speed for tomato plant disease detection. 
\begin{figure}
	\noindent
	\centering
	\includegraphics[width=0.9\linewidth]{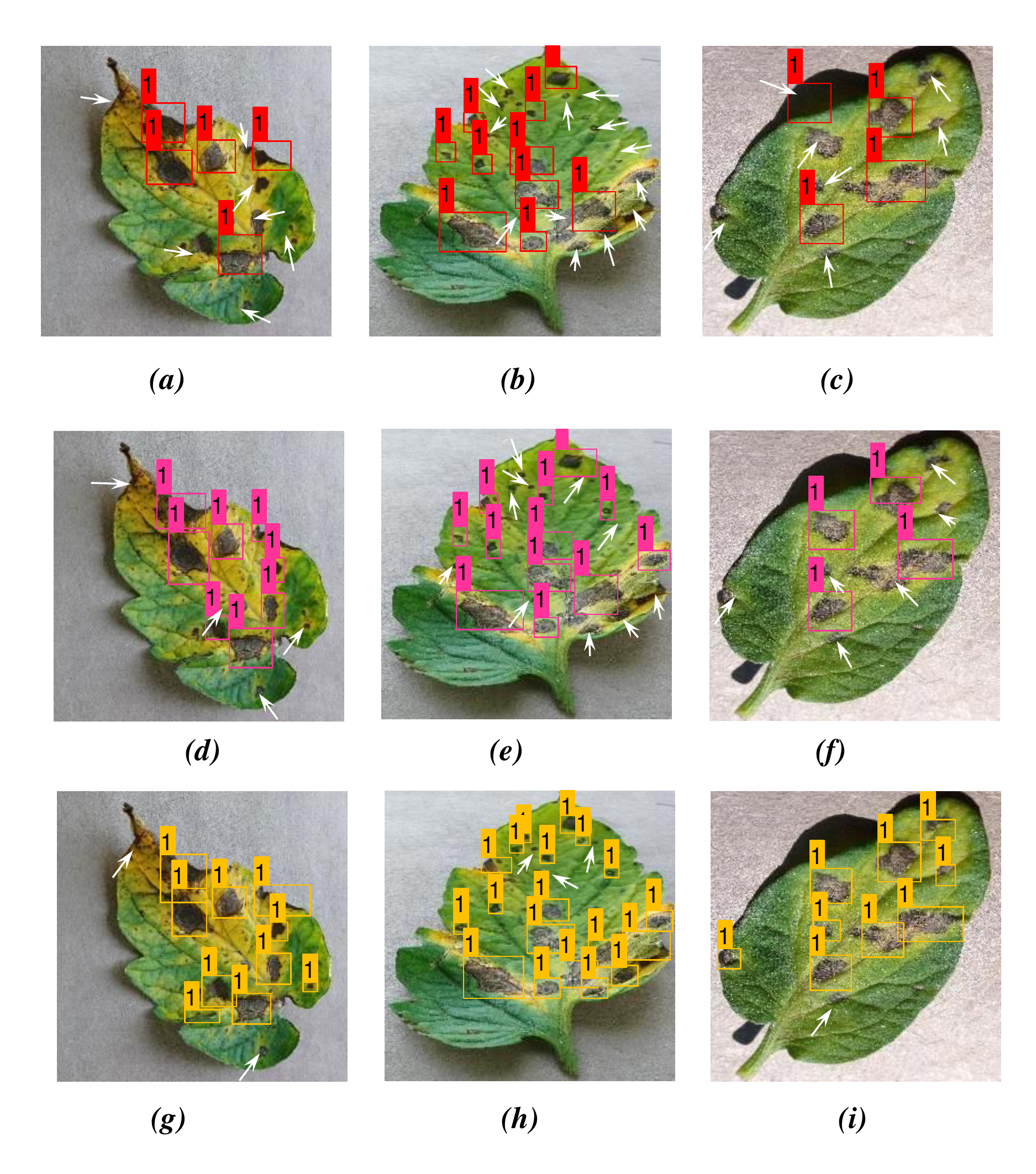}
	\caption{\label{Fig-12} Detection result for early blight on three distinct tomato leaves from three models: (a–c) YOLOv3;  (d–f) YOLOv4; (g-i) proposed model. The white arrow indicates undetected or false detection from the corresponding model prediction.}
\end{figure}
\begin{table}
	\centering
	\caption{Comparison of detection results between YOLOv3, YOLOv4, and proposed model for early blight detection as shown in Fig \ref{Fig-12}. Bold highlights the best result obtained from the corresponding model prediction.}
	\begin{tabular}{c c c c c c  }
		\\[-0.5em]
		\hline
		\\[-0.8em] 
		Figs. No &Model & Detc.   & Undetc.   & Confidence Scores
		\\[-0.0em]
		\hline
		\\[-0.5em]
		Fig. \ref{Fig-12}-(a) & YOLOv3 & 5 & 7  & 0.96, 0.92, 0.84, 0.93, 0.89
		\\
		\\[-0.5em]
		Fig. \ref{Fig-12}-(d) & YOLOv4 & 8 & 4  & \vtop{\hbox{\strut 0.72, 0.90,  0.87, 0.76,  }\hbox{\strut 0.86,  1.00, 0.83, 0.95}}
		\\
		\\[-0.5em]
		\bf{Fig. \ref{Fig-12}-(g)} & \bf{Proposed model} & \bf{10} & \bf{2}  & \vtop{\hbox{\strut 0.97, 0.94,  0.88, 0.96,  1.00, }\hbox{\strut 0.86,  1.00, 0.93, 0.98, 1.00}}
		\\
		\\[-0.5em]
		\hline     
		\\[-0.5em]
		Fig. \ref{Fig-12}-(b) & YOLOv3 & 10 & 12  & \vtop{\hbox{\strut 0.82, 0.93,  0.81, 0.78,  1.00, }\hbox{\strut  1.00, 0.83, 0.95, 0.94, 0.99}}
		\\
		\\[-0.5em]
		Fig. \ref{Fig-12}-(e) & YOLOv4 & 12 & 10  & \vtop{\hbox{\strut0.88, 0.97, 1.00, 1.00, 0.75, 0.86, }\hbox{\strut 0.89, 0.91, 1.00, 1.00, 1.00, 0.96}}    
		\\
		\\[-0.5em]
		Fig. \ref{Fig-12}-(h) & \bf{Proposed model} & \bf{19} & \bf{3}  &  \vtop{\hbox{\strut 0.98, 0.87, 1.0, 1.0, 1.0, 0.86,1.0, 0.97, 0.95, 1.0 }\hbox{\strut 0.99, 0.87, 0.92, 1.0, 1.0, 0.91, 0.98, 0.92, 0.99 }} 
		\\
		\\[-0.5em]
		\hline     
		\\[-0.5em]    
		Fig.\ref{Fig-12}-(c) & YOLOv3 & 4 & 6  & 0.79, 0.88, 0.93, 0.89    
		\\
		\\[-0.5em]
		Fig. \ref{Fig-12}-(f) & YOLOv4 & 4 & 6  & 0.94, 0.98, 1.00,  0.96
		\\
		\\[-0.5em]
		Fig. \ref{Fig-12}-(i) & \bf{Proposed model} & \bf{9} & \bf{1}  & \vtop{\hbox{\strut 0.98, 0.99,  0.98, 0.97,  1.00, }\hbox{\strut 0.96,  1.00, 0.93, 0.98}}            
		\\
		\\[-0.5em]
		\hline
	\end{tabular}
	\label{T-9}
\end{table}

\section{Real-time detection results}
\label{sec:6}

In this section, real-time detection results for the four different disease classes are presented in detail. 
Results include detection of different disease classes, detection in greyscale and in low-resolution images, and at various illumination intensities. 

\subsection{Detection of different plant disease classes}
\label{sec:6a}

\begin{figure}
	\noindent
	\centering
	\includegraphics[width=0.9\linewidth]{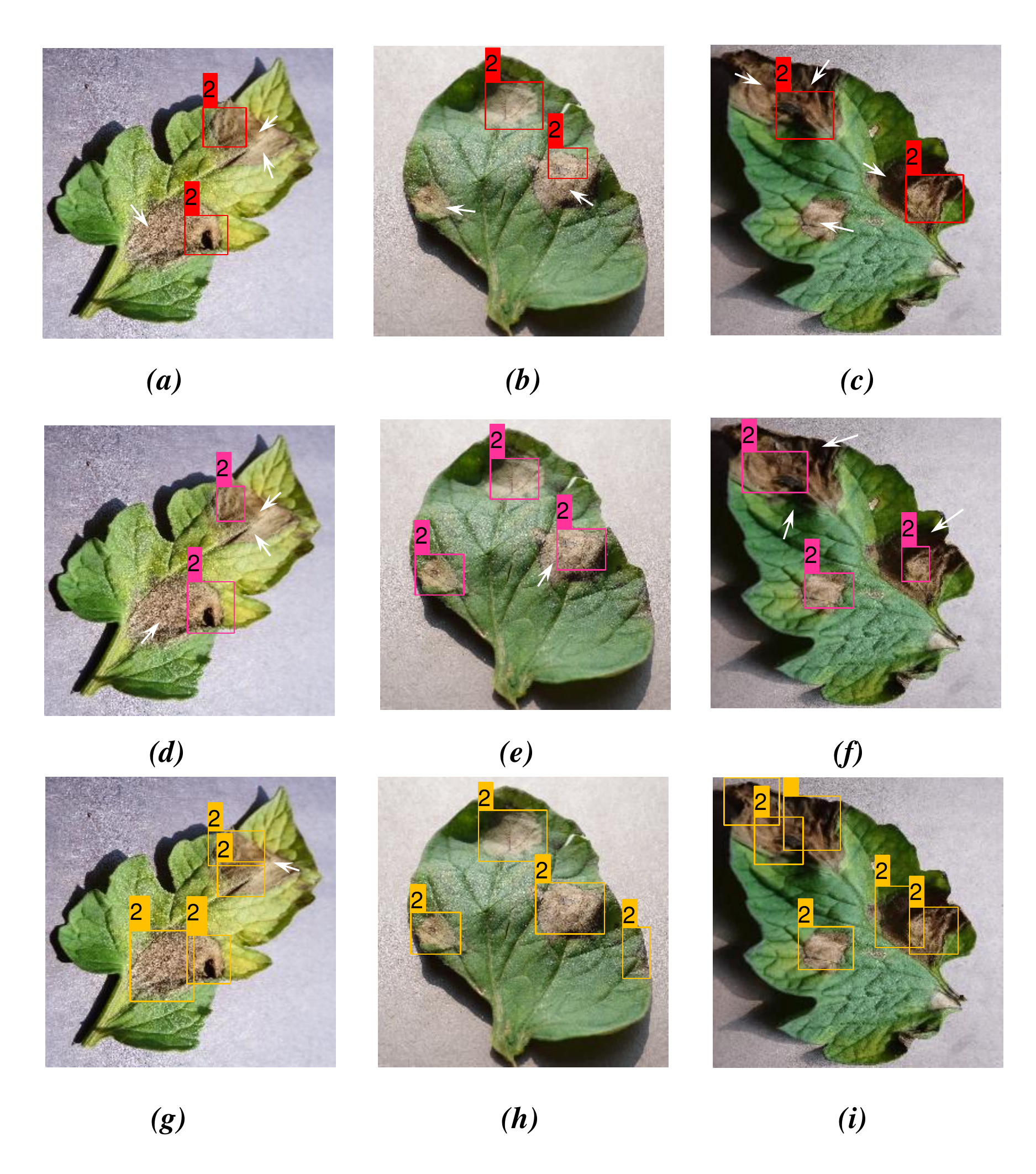}
	\caption{\label{Fig-13} Detection result for late blight on three distinct tomato leaves from three models: (a–c) YOLOv3;  (d–f) YOLOv4; (g-i) proposed model. The white arrow indicates undetected or false detection from the corresponding model prediction.}
\end{figure}

\begin{figure}
	\noindent
	\centering
	\includegraphics[width=0.9\linewidth]{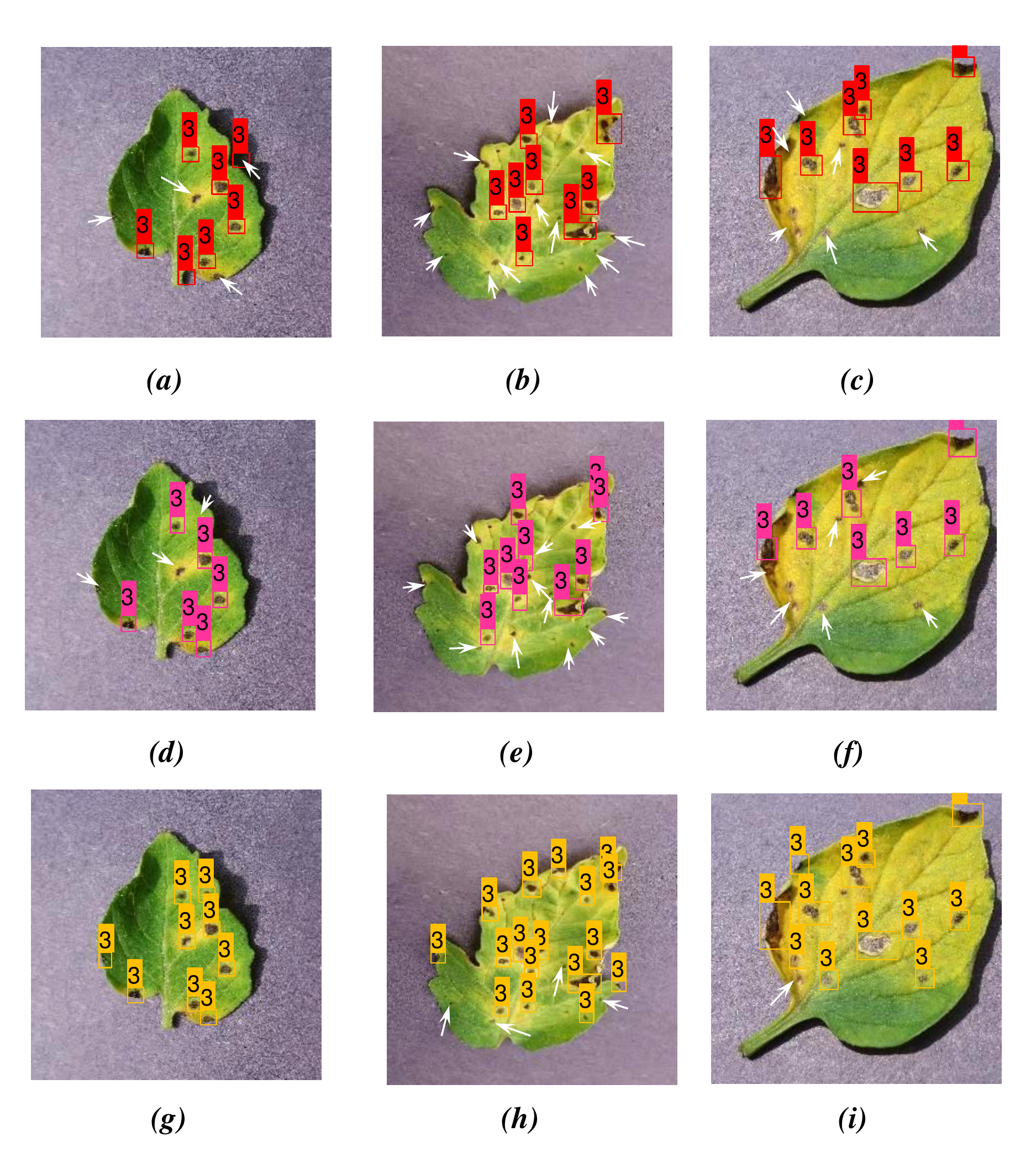}
	\caption{\label{Fig-14} Detection result for Septoria leaf spot on three distinct tomato leaves from three models: (a–c) YOLOv3;  (d–f) YOLOv4; (g-i) proposed model. The white arrow indicates undetected or false detection from the corresponding model prediction.}
\end{figure}

\begin{figure}
	\noindent
	\centering
	\includegraphics[width=0.9\linewidth]{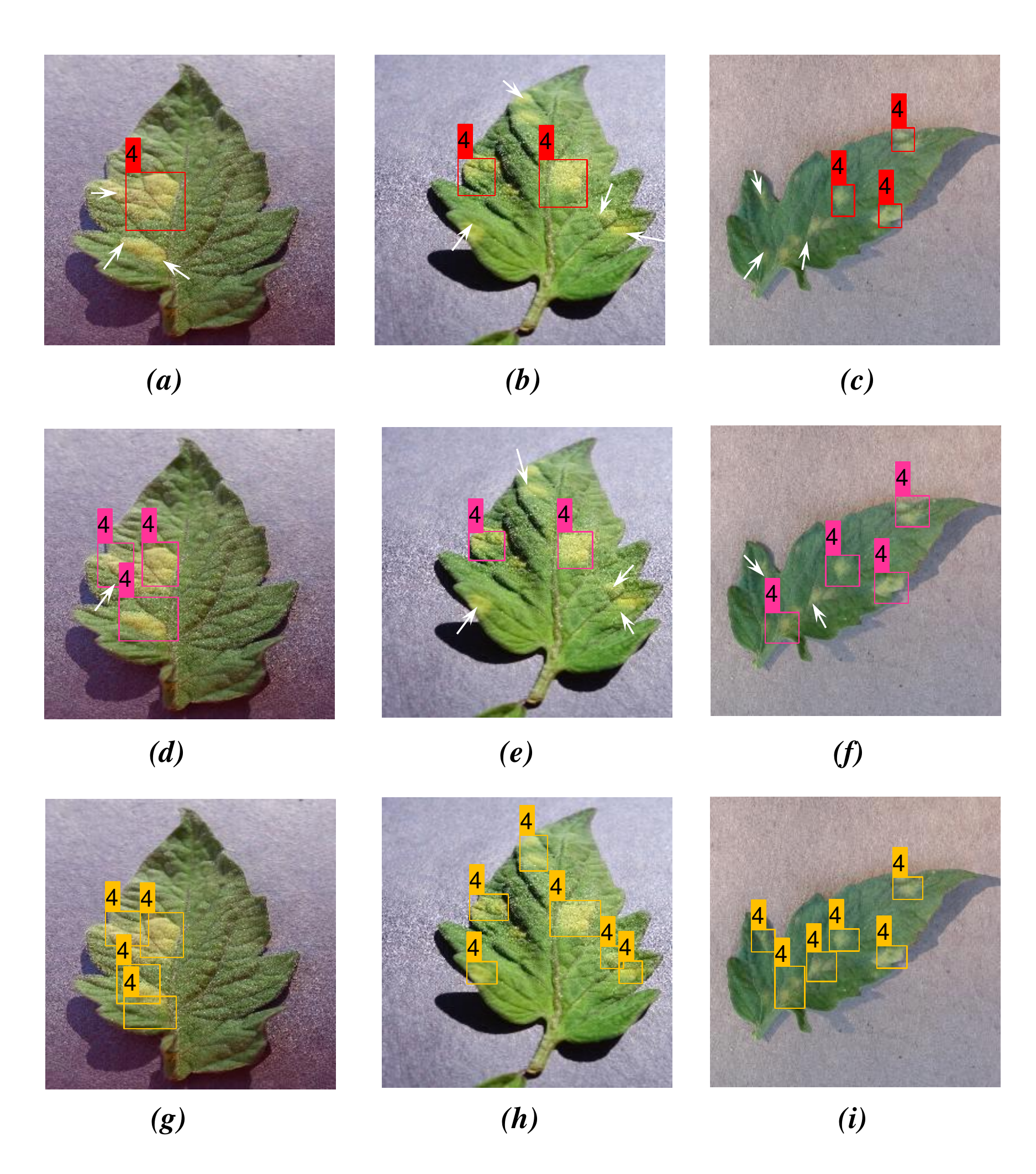}
	\caption{\label{Fig-15} Detection result for leaf mold on three distinct tomato leaves from three models: (a–c) YOLOv3;  (d–f) YOLOv4; (g-i) proposed model. The white arrow indicates undetected or false detection from the corresponding model prediction.}
\end{figure}

Results from the proposed model for four different diseases are compared with predictions from the YOLOv3 and YOLOv4 models. 
Each of these classes are detected in three different leaves. 
Visual representations of the results are presented in Figs. \ref{Fig-12}-\ref{Fig-15}. 
Four diseases: early blight, late blight, Septoria leaf spot, and leaf mold are marked with bounding-box-class identifiers: 1, 2, 3, and 4, respectively. 
The arrows identify undetected objects or false detections by the model. 
\begin{table}
	\centering
	\caption{Comparison of detection results between YOLOv3, YOLOv4, and proposed model for late blight detection  as shown in Fig \ref{Fig-13}. Bold highlights the best result obtained from corresponding model prediction.}
	\begin{tabular}{c c c c c c  }
		\\[-0.5em]
		\hline
		\\[-0.8em] 
		Figs. No &Model & Detc.   & Undetc.   & Confidence Scores
		\\[-0.0em]
		\hline
		\\[-0.5em]
		Fig. \ref{Fig-13}-(a) & YOLOv3 & 2 & 3  &  0.84, 0.93
		\\
		\\[-0.5em]
		Fig. \ref{Fig-13}-(d) & YOLOv4 & 2 & 3  & 0.94, 1.00
		\\
		\\[-0.5em]
		\bf{Fig. \ref{Fig-13}-(g)} & \bf{Proposed model} & \bf{4} & {1}  & 0.98, 1.00, 1.00, 0.97
		\\
		\\[-0.5em]
		\hline     
		\\[-0.5em]
		Fig. \ref{Fig-13}-(b) & YOLOv3 & 2 & 2  & 0.91,  0.94
		\\
		\\[-0.5em]
		Fig. \ref{Fig-13}-(e) & YOLOv4 & 3 & 1  &  0.97, 1.00, 1.00     
		\\
		\\[-0.5em]
		Fig. \ref{Fig-13}-(h) & \bf{Proposed model} & \bf{4} & \bf{0}  &   1.00,  0.96,  1.00, 1.00    
		\\
		\\[-0.5em]
		\hline     
		\\[-0.5em]    
		Fig.\ref{Fig-13}-(c) & YOLOv3 & 2 & 4  & 0.89, 0.87
		\\
		\\[-0.5em]
		Fig. \ref{Fig-13}-(f) & YOLOv4 & 3 & 3  & 0.91, 1.00, 1.00, 
		\\
		\\[-0.5em]
		Fig. \ref{Fig-13}-(i) & \bf{Proposed model} & \bf{6} & \bf{0}  & 0.92, 0.98, 1.00, 1.00, 1.00, 0.97             
		\\
		\\[-0.5em]
		\hline
	\end{tabular}
	\label{T-10}
\end{table}
Detection performance of the three models in detecting each class is compared by reporting detected (detec.) and undetected (undetec.) cases for each leaf in Tables \ref{T-9}-\ref{T-12}. 
Superiority of the proposed model over the original models for the detection task at hand is evident from these results. 

The area of leaves infected by early blight can be identified as mostly dark brown (or black) spots generally distributed discretely as shown in Fig. \ref{Fig-12}. 
After their genesis, early blight lesions grow erratically. 
It is challenging to detect the high-aspect-ratio patches, specifically at the edges of the leaves. 
While YOLOv3 and YOLOV4 miss several early blight spots, the proposed model shows superior performance as shown in Fig. \ref{Fig-12}-(g,h). 
Due to similarity of their textures with the surroundings, these are difficult to detect. 
Even in such tasks, the proposed model provides significant improvement in detection accuracy reflected in reduced number of undetected spots compared to the other two models, visually illustrated in  Figs. \ref{Fig-12} -(b), (e) and (h), and also in results tabulated in Table \ref{T-9}. 

\begin{table}
	\centering
	\caption{Comparison of detection results between YOLOv3, YOLOv4, and proposed model for Septoria leaf spot detection  as shown in Fig \ref{Fig-14}. Bold highlights the best result obtained from corresponding model prediction.}
	\begin{tabular}{c c c c c c  }
		\\[-0.5em]
		\hline
		\\[-0.8em] 
		Figs. No &Model & Detc.   & Undetc.   & Confidence Scores
		\\[-0.0em]
		\hline
		\\[-0.5em]
		Fig. \ref{Fig-14}-(a) & YOLOv3 & 6 & 3  & 0.96, 0.92, 1.00, 0.84, 0.93, 0.89
		\\
		\\[-0.5em]
		Fig. \ref{Fig-14}-(d) & YOLOv4 & 6 & 3  & 0.94, 1.00,  0.97, 0.86,  1.00, 0.87
		\\
		\\[-0.5em]
		\bf{Fig. \ref{Fig-14}-(g)} & \bf{Proposed model} & \bf{9} & \bf{0}    & \vtop{\hbox{\strut  {1.00, 1.00, 1.00, 0.99, 0.87}}\hbox{\strut {1.00,  0.91, 0.92, 1.00 }}}             
		\\
		\\[-0.5em]
		\hline     
		\\[-0.5em]
		Fig. \ref{Fig-14}-(b) & YOLOv3 & 8 & 13  & \vtop{\hbox{\strut  1.00, 1.00, 1.00, 0.86}\hbox{\strut 0.99, 0.97, 0.87,  0.91 }}
		\\
		\\[-0.5em]
		Fig. \ref{Fig-14}-(e) & YOLOv4 & 10 & 11  & \vtop{\hbox{\strut 0.99, 0.87, 1.00, 1.00, 1.00,}\hbox{\strut 0.97, 0.99, 0.84,  0.91, 1.00 }}
		\\
		\\[-0.5em]
		Fig. \ref{Fig-14}-(h) & \bf{Proposed model} & \bf{17} & \bf{4}  & \vtop{\hbox{\strut 0.94, 0.97, 1.00, 1.00, 1.00, 0.86}\hbox{\strut 1.00, 0.94, 0.99, 0.99, 0.97, 0.94}\hbox{\strut 1.00, 1.00, 0.91, 0.98, 0.99 }}
		\\
		\\[-0.5em]
		\hline     
		\\[-0.5em]    
		Fig.\ref{Fig-14}-(c) & YOLOv3 & 7 & 6  & 0.79, 0.87, 0.91, 0.83, 1.00, 1.00, 0.94     
		\\
		\\[-0.5em]
		Fig. \ref{Fig-14}-(f) & YOLOv4 & 7 & 6  & 0.89, 0.91, 1.00, 1.00, 1.00, 0.96
		\\
		\\[-0.5em]
		Fig. \ref{Fig-14}-(i) & \bf{Proposed model} & \bf{12} & \bf{1}  & \vtop{\hbox{\strut 0.94, 0.97, 1.00, 1.00, 1.00, 0.99,}\hbox{\strut 1.00, 1.00, 0.91, 0.92, 0.99, 1.00 }}             
		\\
		\\[-0.5em]
		\hline
	\end{tabular}
	\label{T-11}
\end{table}

The late blight infections usually show up as larger dark brown lesions in an arbitrary area of a tomato plant which can rapidly spread. 
Due to irregular shapes and similarity of their texture with surrounding leaves, it is often hard to detect precisely. 
While the detection results from the YOLOv3 and YOLOv4 show several missed detections when the interface between infected zones and leaves are not prominent as shown in Fig.\ref{Fig-13} -(a-f), the proposed model yields much better performance in such challenging scenarios by identifying coalescent infected zones between multiple infected spots in Fig.\ref{Fig-13} -(g-i). 
Again the proposed model is able to reduce missed detections. 
Moreover, it provides higher confidence scores in bounding box predictions compared to the other two models as shown in Table \ref{T-10}. 
\begin{table}
	\centering
	\caption{Comparison of detection results between YOLOv3, YOLOv4, and proposed model for leaf mold detection  as shown in Fig \ref{Fig-15}. Bold highlights the best result obtained from corresponding model prediction.}
	\begin{tabular}{c c c c c c  }
		\\[-0.5em]
		\hline
		\\[-0.8em] 
		Figs. No &Model & Detc.   & Undetc.   & Confidence Scores
		\\[-0.0em]
		\hline
		\\[-0.5em]
		Fig. \ref{Fig-15}-(a) & YOLOv3 & 1 & 3  & 0.96
		\\
		\\[-0.5em]
		Fig. \ref{Fig-15}-(d) & YOLOv4 & 3 & 1  & 0.94, 1.00,  0.97
		\\
		\\[-0.5em]
		\bf{Fig. \ref{Fig-15}-(g)} & \bf{Proposed model} & \bf{4} & \bf{0}  & 0.91, 1.00, 1.00, 0.94
		\\
		\\[-0.5em]
		\hline     
		\\[-0.5em]
		Fig. \ref{Fig-15}-(b) & YOLOv3 & 2 & 4  &   0.94, 0.96
		\\
		\\[-0.5em]
		Fig. \ref{Fig-15}-(e) & YOLOv4 & 2 & 4  &  0.97, 1.00
		\\
		\\[-0.5em]
		Fig. \ref{Fig-15}-(h) & \bf{Proposed model} & \bf{6} & \bf{0}  &  0.91, 1.00,  0.96, 0.89,  1.00, 1.00
		\\
		\\[-0.5em]
		\hline     
		\\[-0.5em]    
		Fig.\ref{Fig-15}-(c) & YOLOv3 & 3 & 3 &  0.93, 0.89, 1.00     
		\\
		\\[-0.5em]
		Fig. \ref{Fig-15}-(f) & YOLOv4 & 4 & 2  & 0.89, 0.91, 1.00, 1.00
		\\
		\\[-0.5em]
		Fig. \ref{Fig-15}-(i) & \bf{Proposed model} & \bf{6} & \bf{0}  & 0.89, 0.91, 1.00, 1.00, 1.00, 0.96             
		\\
		\\[-0.5em]
		\hline
	\end{tabular}
	\label{T-12}
\end{table}

Septoria leaf spots have visual similarities to the early blight (Fig. \ref{Fig-14}). 
However, these spots are darker and more rounded. 
For relatively less dense discrete distribution of Septoria spots, all three models perform reasonably well as shown in Fig. \ref{Fig-14} -(a, d, g). 
Several missed detections for relatively densely distributed spots are obtained for YOLOv3 and YOLOv4 in Fig. \ref{Fig-14} -(c,f). 
On the other hand, the proposed model reduces missed detections significantly as shown in Fig. \ref{Fig-14} -(i). 
In more challenging scenarios such as densely populated distribution of infected areas, as in Fig. \ref{Fig-14} -(h), the proposed model illustrates its superiority, correctly predicting higher confidence scores for bounding boxes along with significant reduction in missed detections as listed in Table \ref{T-11}. 

Lastly, detection results for leaf-mold-infected leaves are presented in Fig. \ref{Fig-15}. 
Leaf molds are generally pale greenish-yellow or olive-green spots. 
The associated difficulty in detection is due to color similarity and also indistinguishable margins with surrounding leaves. 
Varying lightening conditions in the image dataset is an additional difficulty for the object detection models to detect each of the lesions precisely. 
The proposed model is again superior to the YOLOv3 and YOLOv4 as shown in Fig. \ref{Fig-15}-(h,i), in particular, closely distributed Septoria lesions. 
For the proposed model, miss detections are reduced the confidence scores are improved, is evident from Table \ref{T-12}. 

Based on the detection results for the four classes, overall, the proposed model illustrates superior detection ability, in particular, in detecting densely populated fine-grain diseases, irregular geometric morphology of the infected areas, the coexistence of multi-scale infection lesions, similarity of textures of affected areas and surroundings, varying lightening conditions, etc. compared to the original YOLO models. 

\subsection{Detection under greyscale and  low-resolution images}
\label{sec:6b}

\begin{figure}
	\noindent
	\centering
	\includegraphics[width=0.9\linewidth]{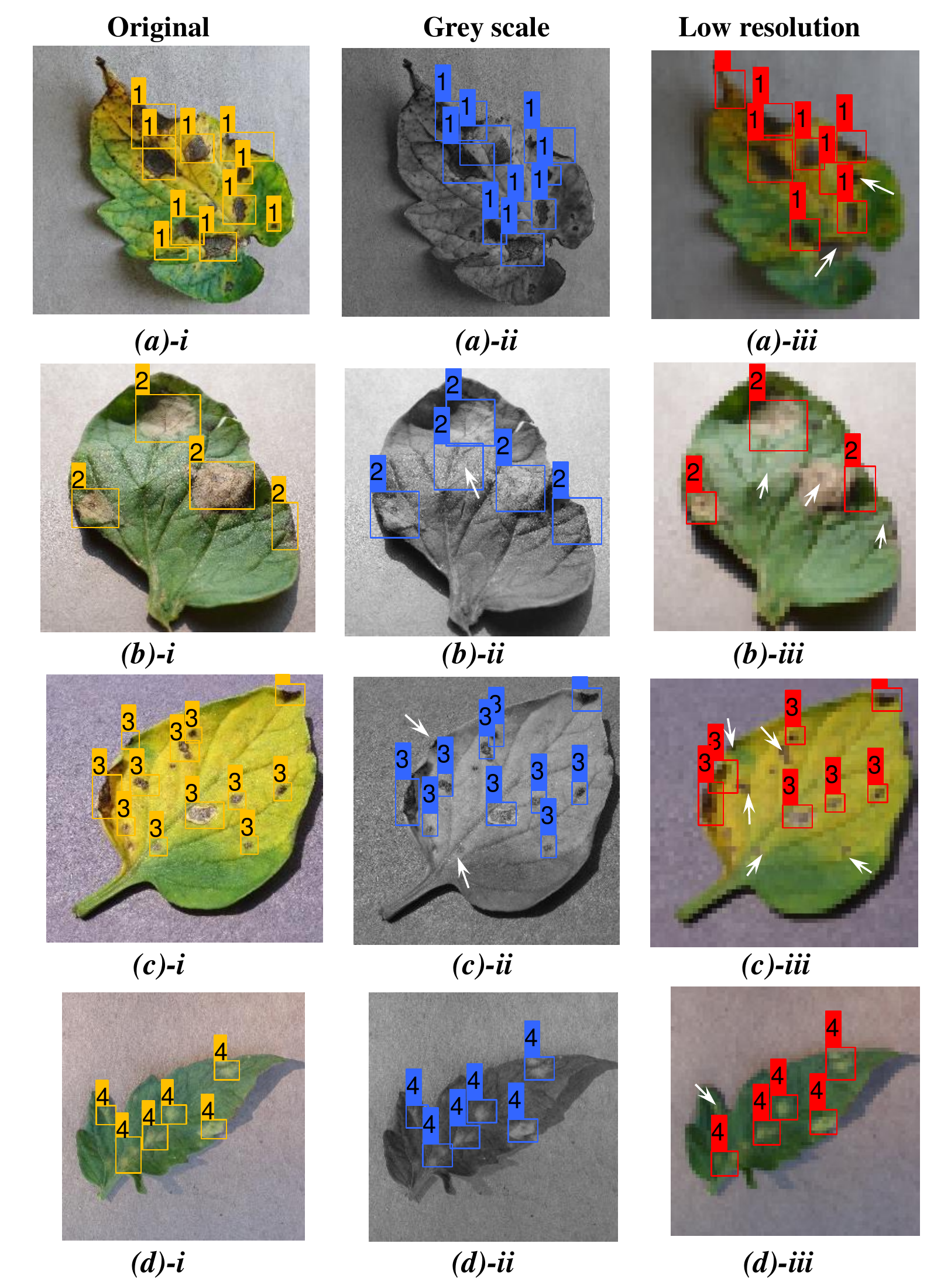}
	\caption{\label{Fig-16} Detection  results on  original RGB, corresponding  greyscale, and pixelated low-resolution images for [(a)-i, ii, iii]  early blight; [(b)-i, ii, iii] late blight; [(c)-i, ii, iii] Septoria; and  [(d)-i, ii, iii] leaf mold from the proposed model. White arrow  indicates undetected or false detection from the model prediction.}
\end{figure}

The proposed model is used to predict greyscale and pixelated low-resolution $(100\times100)$ images. 
The predictions are then compared to the original red, green, and blue (RGB) images to analyze the detection accuracy of the bounding boxes for all four disease classes in Fig \ref{Fig-16}. 

%
\begin{table}
	\centering
	\caption{Comparison of detection  results on the  original RGB, corresponding  greyscale, and pixelated low resolution  images from the proposed model as shown in Fig. \ref{Fig-16}.}
	\begin{tabular}{c*{8}{c}}
        \hline
		{ Figs. No} & \vtop{\hbox{\strut Disease }\hbox{\strut class}} &
		\multicolumn{2}{c}{ Original } &
		\multicolumn{2}{c}{Grey scale} &
		\multicolumn{2}{c}{ Low resolution } \\
		& & Detec.   & Undetec.  & Detec.   & Undetec. & Detec.   & Undetec. \\ \hline  \\
		Figs. \ref{Fig-16}-(a)-i, ii, iii   & Early blight   & 10    & 2    & 10    & 2   & 8     & 4       \\
		Figs. \ref{Fig-16}-(b)-i, ii, iii   & Late blight   & 4    & 0    & 4    & 0   & 2     & 2       \\ 
		Figs. \ref{Fig-16}-(c)-i, ii, iii   & Septoria   & 12    & 1    & 10    & 3   & 7     &6       \\ 
		Figs. \ref{Fig-16}-(d)-i, ii, iii   & Leaf mold   & 6   & 0   & 6    & 0   & 5     & 1       \\ 
		\\ \hline 
	\end{tabular}
	\label{T-13}
\end{table}
The proposed model is highly accurate in predicting greyscale images, for early blight and leaf mold classes in particular, with zero missed detections. 
However, the model misses one spot for late blight and two spots for Septoria as shown in Fig \ref{Fig-16} -(b-ii, c-ii). 
These results indicate that the accuracy of the detection model can reduce for low-resolution greyscale images with some undetected disease spots in densely populated areas and similar textural colored backgrounds. 
Nevertheless, detection accuracy of the proposed model for low-resolution images is higher than the YOLOv3 and YOLOv4 models (see Tables \ref{T-9}- \ref{T-12} and Table \ref{T-13}). 
It can be concluded from the test results that the proposed model is more adaptive in more challenging environments compared to the original models. 
\begin{figure}
	\noindent
	\centering
	\includegraphics[width=0.9\linewidth]{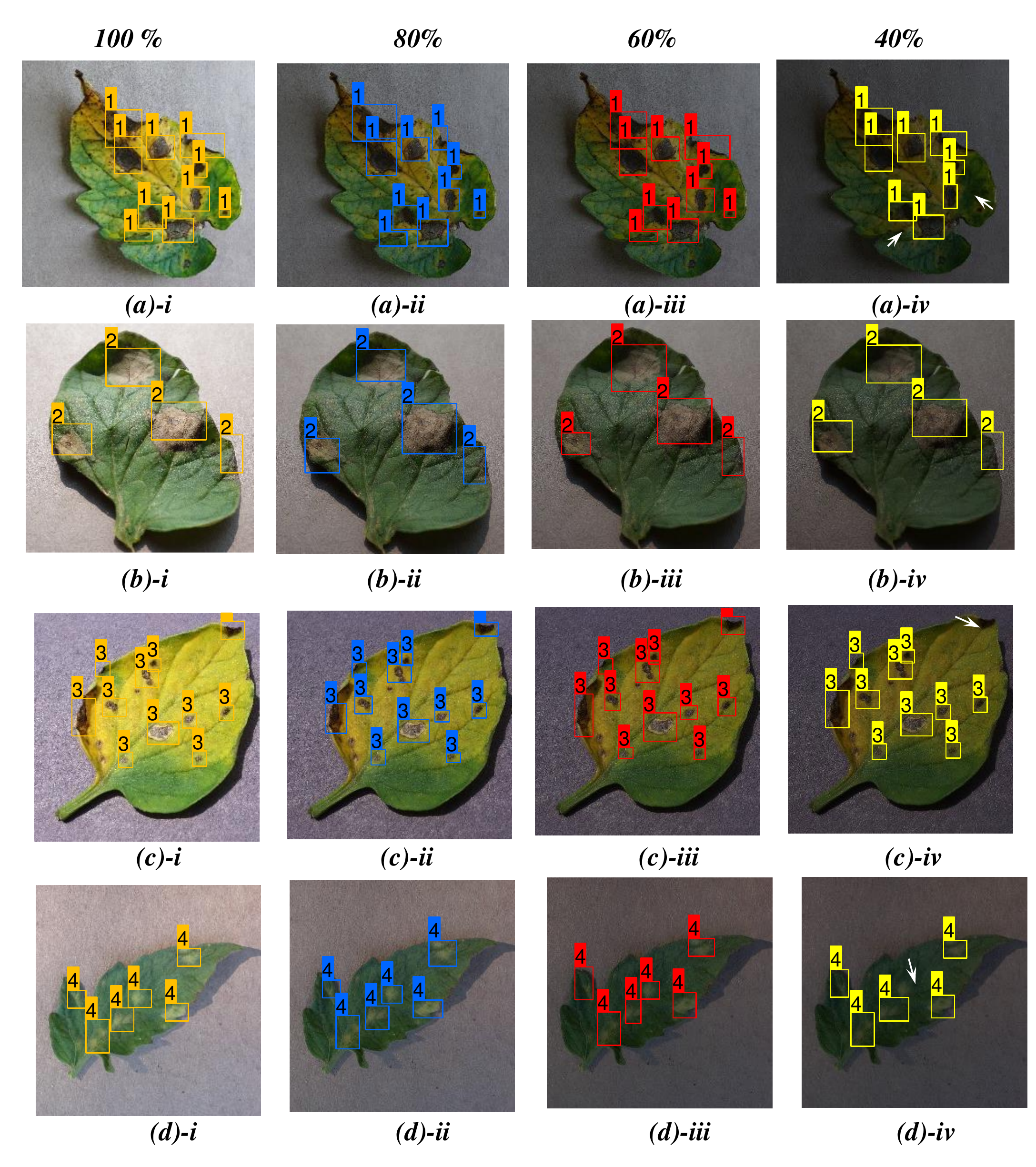}
	\caption{\label{Fig-17} Detection  results obtained from the   original RGB and different illumination conditions including  $90\%$, $80\%$, and $60\%$  brightness intensity  for   [(a)-i, ii, iii, iv]  early blight; [(b)-i, ii, iii, iv] late blight; [(c)-i, ii, iii, iv] Septoria; and  [(d)-i, ii, iii, iv] leaf mold from the proposed model. White arrow  indicates undetected or false detection from the model prediction.}
\end{figure}
\begin{table}
	\centering
	\caption{Comparison of detection results obtained from the   original RGB and different illumination conditions including  $90\%$, $80\%$, and $60\%$  brightness intensities  as shown in Fig. \ref{Fig-17}.}
	\begin{tabular}{c*{10}{c}}
        \hline
		{ Figs. No} & \vtop{\hbox{\strut Disease }\hbox{\strut class}} &
		\multicolumn{2}{c}{ Original} &
		\multicolumn{2}{c}{90$\%$} &
		\multicolumn{2}{c}{ 80$\%$} & 
		\multicolumn{2}{c}{ 60$\%$} \\
		& & Det.   & Undet.  & Det.   & Undet. & Det.   & Undet. & Det.   & Undet. \\ \hline \\ 
		Figs. \ref{Fig-17}-(a)i-iv   & Early blight   & 10    & 2    & 10    & 2   & 10     & 2   & 8     & 4      \\
		Figs. \ref{Fig-17}-(b)i-iv   & Late blight   & 4    & 0    & 4    & 0   & 4     & 0    & 3     & 1   \\ 
		Figs. \ref{Fig-17}-(c)i-iv   & Septoria   & 12    & 1    & 12    & 1   & 10     & 3   & 9     & 4   \\ 
		Figs. \ref{Fig-17}-(d)i-iv   & Leaf mold   & 6   & 0   & 6    & 0   & 5     & 1  & 5     & 1    \\ 
		\\ \hline 
	\end{tabular}
	\label{T-14}
\end{table}

\subsection{Detection  under different illumination intensities}
\label{sec:6c}

In Fig. \ref{Fig-17}, prediction results from the proposed model under different illumination intensities are analyzed. 
Detection results for three different brightness conditions: $80\%$, $60\%$ , and $40\%$ are compared to the original detection results under normal brightness conditions. 
Here, white arrow indicates undetected spots or false detections compared to the original images. 
One can see that the model can accurately detect disease spots in different brightness conditions, in particular, $80\%$ and  $60\%$ brightness conditions. 
However, missed and false detections are obtained for early blight, Septoria,  and leaf mold in $40\%$ brightness intensity as shown in Fig. \ref{Fig-17}-(a, c, d). 
Nevertheless, predictions of boundary boxes and corresponding confidence scores (see Table \ref{T-14}) for the proposed model are of reasonable accuracy under dimming conditions which justifies the use of the model in practical on-field disease detection tasks. 

Despite its efficacy for the tomato plant disease detection task, there is scope for improvement in the proposed model's feature extraction process to further minimize missed and false detections under challenging backgrounds. 
Future work will focus on further optimization of detection speed and accuracy of the model for use in a mobile computing platform for portable on-field detection. 

\section{Conclusion}
\label{sec:7}

In this work, a real-time object detection model is developed based on the YOLOv4 algorithm. 
Several modifications are proposed to optimize both detection accuracy and speed of the model which are then verified in various complex detection tasks in noisy environments utilizing traditional performance measures for detecting objects. 
Accuracy, efficiency and robustness of the model are enhanced due to introduction of the CSP1-$n$ block in the backbone that improved feature extraction and the CSP2-$n$ module in the neck that preserved fine-grain local information. 
Moreover, a DenseNet block is implemented in the backbone bettering feature transfer and reuse. 
Despite use of these additional components, the model is demonstratively easier to train than the original YOLO models. 
Additionally, use of the Hard-Swish as the primary activation function ameliorates the model's learning capability of nonlinear characteristics in image features. 
All these modifications are able to increase the accuracy of the model substantially. 
The current algorithm achieves the highest detection accuracy and speed compared to some existing state-of-the-art detection models. 
At a detection rate of 70.19 $FPS$, the proposed model yielded a precision value of $90.33 \%$, $F1-$score of $93.64 \%$, $mAP$ of $96.29 \%$. 
Compared to the original YOLOv4, the proposed model acquires $8.94\%$ increase in precision,  $7.20\%$ increase in the $F1-$score, and provides $10.34\%$ faster detection illustrating the superior potential of the model in real-time in-field applications. 
Current work provides an efficient method for accurate fine-grain detection of different object classes in complex scenarios. 
Although the model is only applied for plant disease detection herein, it can be extended to different fruit and crop detection tasks, generic disease detection problems, and various automated agricultural detection processes \cite{shamshirband2019survey,fan2020spatiotemporal}.

\bibliographystyle{spphys}       
\bibliography{reference}

\end{document}